\documentclass{opt2021} 
\usepackage{times}

\usepackage{amsmath,amsfonts,bm}

\def\eqref#1{equation~\ref{#1}}

\def\1{\bm{1}}

\def\mE{{\bm{E}}}

\def\mU{{\bm{U}}}
\def\mV{{\bm{V}}}
\def\mW{{\bm{W}}}

\DeclareMathAlphabet{\mathsfit}{\encodingdefault}{\sfdefault}{m}{sl}
\SetMathAlphabet{\mathsfit}{bold}{\encodingdefault}{\sfdefault}{bx}{n}
\newcommand{\tens}[1]{\bm{\mathsfit{#1}}}

\def\tW{{\tens{W}}}

\def\gG{{\mathcal{G}}}

\def\gZ{{\mathcal{Z}}}

\newcommand{\R}{\mathbb{R}}

\DeclareMathOperator*{\argmin}{arg\,min}

\usepackage{hyperref}
\usepackage{url}
\usepackage{graphicx} 
\usepackage{xcolor}
\usepackage{caption}
\usepackage{mathtools}
\usepackage{amsmath}
\usepackage{physics}
\usepackage{comment}
\usepackage{siunitx}
\usepackage{algorithm}
\usepackage[noend]{algorithmic}
\usepackage{lscape}
\usepackage{multirow}
\usepackage{multicol}
\usepackage{pdflscape}
\usepackage{wrapfig}
\usepackage{soul}
\usepackage{pifont}

\graphicspath{ {./figures/} }

\newcommand{\GlobalLR}{\ensuremath{\eta}}

\newcommand{\Parameters}{\ensuremath{\theta}}

\newcommand{\HyperParameters}{\ensuremath{\lambda}}

\newcommand{\ZPerc}{\ensuremath{\mathcal{Z}}}

\newcommand{\AlgName}{autoHyper}

\makeatletter
\def\hlinewd#1{%
  \noalign{\ifnum0=`}\fi\hrule \@height #1 \futurelet
   \reserved@a\@xhline}

\definecolor{custom_green}{RGB}{40, 100, 219}
\definecolor{changed}{RGB}{0, 0, 0}
\definecolor{red}{RGB}{255, 100, 100}
\definecolor{green}{RGB}{40, 219, 100}
\definecolor{best}{RGB}{10, 150, 10}
\definecolor{close}{RGB}{255, 140, 0}

\title{Towards Robust and Automatic Hyper-Parameter Tunning}

\optauthor{%
\Name{Mathieu Tuli} \Email{mathieutuli@cs.toronto.edu}\\
\addr University of Toronto, Canada\\
\Name{Mahdi S. Hosseini} \Email{mahdi.hosseini@unb.ca}\\
\addr University of New Brunswick, Canada\\
\Name{Konstantinos N. Plataniotis} \Email{kostas@ece.utoronto.ca}\\
\addr University of Toronto, Canada
}

\begin{document}
\maketitle
\vspace{-10mm}
\begin{abstract}
The task of hyper-parameter optimization (HPO) is burdened with heavy computational costs due to the intractability of optimizing both a model's weights and its hyper-parameters simultaneously. In this work, we introduce a new class of HPO method and explore how the low-rank factorization of the convolutional weights of intermediate layers of a convolutional neural network can be used to define an analytical response surface \cite{bergstra2012random} for optimizing hyper-parameters, using only training data. We quantify how this surface behaves as a surrogate to model performance and can be solved using a trust-region search algorithm, which we call \AlgName. The algorithm outperforms state-of-the-art such as Bayesian Optimization and generalizes across model, optimizer, and dataset selection. Our code can be found at \url{https://github.com/MathieuTuli/autoHyper}.
\end{abstract}

\section{Introduction}
\begin{wrapfigure}{r}{0.4\textwidth}
\vspace{-8mm}
\begin{center}
\includegraphics[width=.4\textwidth]{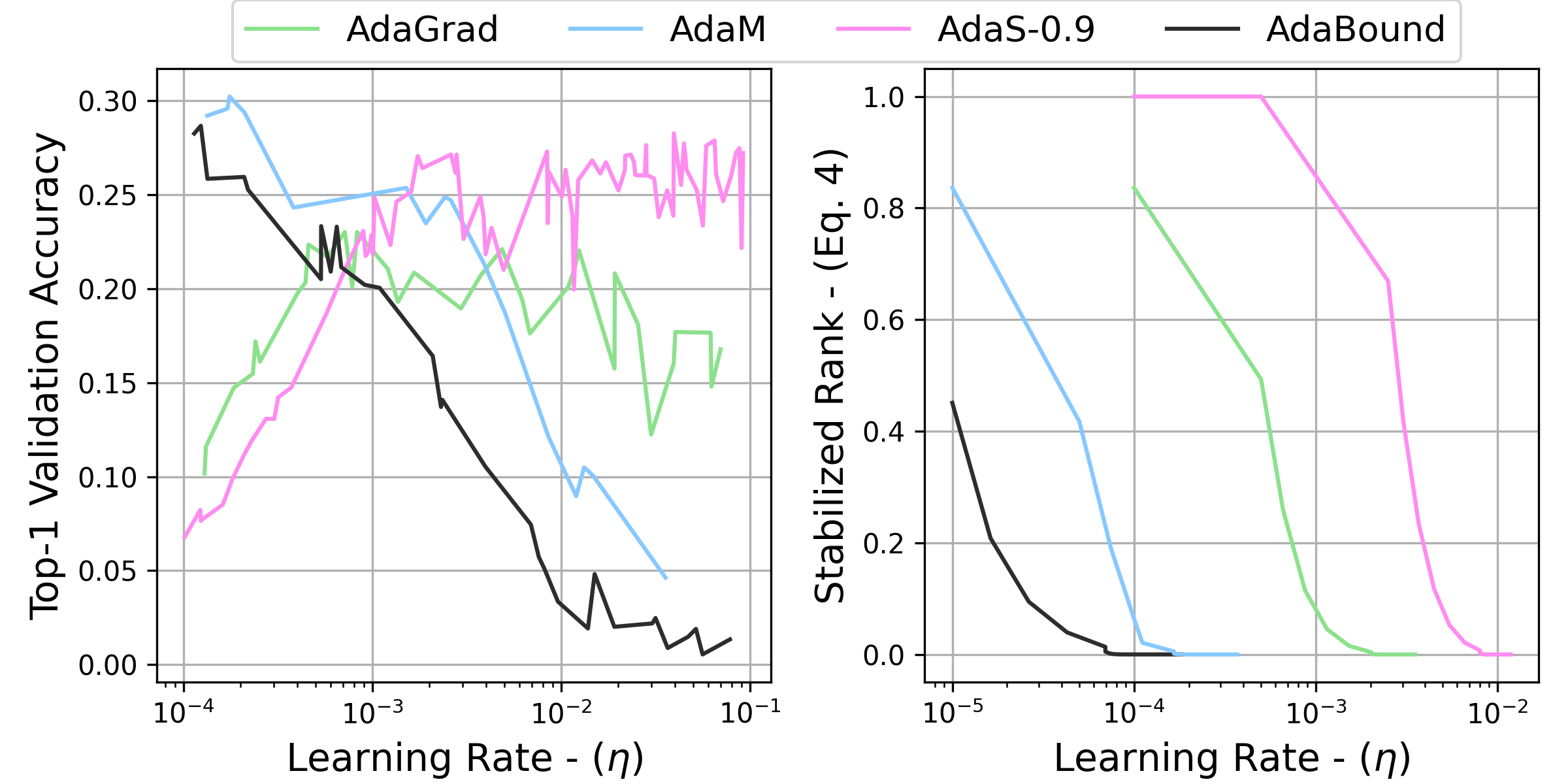}
\end{center}
\vspace{-5mm}
\caption{Left: Validation Accuracy vs. Right: stable rank metric tracked \AlgName\ (see \autoref{sec:kg})}
\label{fig:zeta_comp}
\vspace{-6mm}
\end{wrapfigure}

The task of hyper-parameter optimization (HPO) is burdened with computational intractability caused by a dual-optimization problem, whereby optimization over a network's weights as well as its hyper-parameters cannot happen simultaneously  \cite{bergstra2012random}. The abstract formulation of HPO can be defined as \\$\HyperParameters^* \leftarrow \argmin\limits_{\HyperParameters \in \Lambda}\{\mathbb{E}_{x \sim M}[
    \mathcal{L}(x;\mathcal{A}_{\HyperParameters}({X}^{(\text{train})})]\}$ as defined by \citet{bergstra2012random},
where $X^{(\text{train})}$ and $x$ are random variables, modelled by some natural distribution $M$, that represent the train and validation data, respectively. {\color{changed}$\mathcal{L}(\cdot)$ is some expected loss and $\mathcal{A}_{\HyperParameters}({X}^{(\text{train})})$ is a learning algorithm that maps ${X}^{(\text{train})}$ to some learned function, conditioned on the hyper-parameter set $\HyperParameters$. Note that this learned function, denoted as $f(\Parameters;\HyperParameters;X^{(\text{train})})$, involves its own independent inner optimization problem}. Because of this, optimization over the hyper-parameters  $\HyperParameters$ cannot occur until optimization over $f(\Parameters;\HyperParameters;X^{(\text{train})})$ is complete. This means that HPO in this form suffers from heavy computational burden and is practically unsolvable. However, \citet{bergstra2012random} showed that this burden is reduced if we simplify our scope and only consider $\HyperParameters^* \leftarrow \argmin\limits_{\HyperParameters \in \Lambda}\tau(\HyperParameters)$
where, $\tau$ is called the \textit{response surface} and $\Lambda$ is some set of choices for $\HyperParameters$ (\textit{i.e. the search space}). Simply put, the goal of the response surface is to act as an easier to solve surrogate function parameterized by $\HyperParameters$ whose minimization is correlated to the minimization of our networks's objective function. Importantly, this response surface is supposed to be much easier to solve for, and removes the dual-optimization problem.

Unfortunately, little advancements in an analytical model of the response surface $\tau$ has led to estimating it by (a) running multiple trials of different HP configurations (\textit{e.g. Random Search}) against validation datassets; or (b) characterizing the distribution model of a certain configuration's performance metric (\textit{e.g. validation performances}) to numerically define a relationship between $\tau$ and $\HyperParameters$ (\textit{e.g. Bayesian Optimization}). Despite the success of these methods, they exist as estimations of a response surface, which is itself already a simplification/estimation of our initial objective function, which we argue is inefficient and sub optimal. We work towards resolving this issue.

In this paper, we deviate from existing classes of HPO methods and explore an alternative surrogate metric that demonstrates how to perform almost fully automatic HPO using only the training dataset. Our contributions are as follows: \textbf{(1)} stemming from the notion of \textit{stable rank} in \citep{hosseini2020adas, jaegerman2021search}, we introduce the task of monitoring the well-posedness of learning layers in a Convolutional Neural Network (CNN) in order to develop a well-defined analytical response surface. Figure \ref{fig:zeta_comp} shows how our new metric behaves well and tractably in contrast to conventional validation performance measures. This response surface deviates from existing works and exists as a new class of HPO; \textbf{(2)} we propose a trust-region search algorithm, dubbed \AlgName, to optimize our response surface and conduct HPO using only the training set. This algorithm almost eliminates all need for human intuition or manual intervention, and is not bound by a manually set searching space, paving the way towards automatic HPO; and \textbf{(3)}, we extend the \AlgName\ algorithm to multi-dimensional HPO.

\section{A New Response Surface Model}\label{sec:kg}
 
\subsection{Stable Rank via Low-Rank Factorization}
We wish to analyze the weight matrices of our neural network and develop a metric that we can track, per epoch, that acts as a surrogate to validation performance. To do so, we decompose the weight matrices of the network and study them by use of low-rank factorization, similar to what \citet{hosseini2021conet} did for channel size optimization. Consider the 4-D tensor $\tW \in \R^{N_1 \times N_2 \times N_3 \times N_4}$ as the weights of a layer in a CNN ($N_1$ \& $N_2$ being the height and width of kernel size, $N_3$ \& $N_4$ the input and output channel size, respectively). We decompose $\tW$ along some dimension $d$ as
\begin{align*}
    \tW [\texttt{4-D Tensor}] \xrightarrow{\text{unfold}}
    \mW_d [\texttt{2-D Matrix}] \xrightarrow{\text{factorize + decompose}} 
    \widehat{\mU}_d \widehat{\Sigma}_d \widehat{\mV}_d^T + \mE_d.
\end{align*}
\begin{wrapfigure}{r}{0.4\textwidth}
    \vspace{-4mm}
    \centering
    \includegraphics[width=0.4\textwidth]{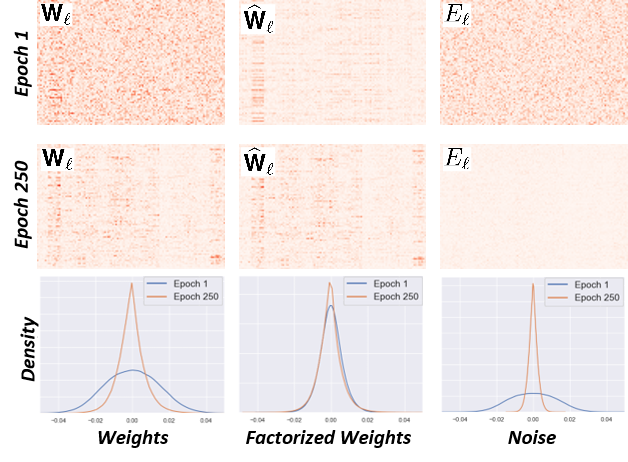}
    \vspace{-4mm}
    \caption{Low-Rank Decomposition (figure replicated from \cite{hosseini2021conet})}
    \label{fig:lowrank_concept}
    \vspace{-4mm}
\end{wrapfigure}
We subsequently define $\widehat{\mW}_d=\widehat{\mU}_d \widehat{\Sigma}_d \widehat{\mV}_d^T$ for simplicity, where $\widehat{\mW}_d$ is the low-rank matrix containing limited non-zero singular values. We use the Variational Bayesian Matrix Factorization (VBMF) \citep{nakajima2013global} for the low-rank factorization. This factorization is critical as it captures the presence of noise, allowing analysis to be invariant to the randomness of initialization. Note, unfolding isn't necessary for linear layers (e.g. for LSTMs), and the analysis is the same. Through this factorization, initially, the low-rank component $\widehat{\mW}_d$ has empty structure (i.e. $\widehat{\mW}_d=\oslash$) as the randomness of the initialized weights is fully captured by $\mE_d$. As training progresses, the low-rank component will gain structure and becomes non-empty as the network learns to map inputs to outputs. This is visualized in Figure \ref{fig:lowrank_concept} for a ResNet34 layer. Notice how the low-rank matrix maintains its structure and strengthens its structure over training, while the perturbing noise element decays. We state that such behaviour leads to a stabilized encoding layer and indicates a beneficial progress in learning. Following \cite{hosseini2020adas}, we define the \textit{stable rank} of the weight matrix as
\begin{equation}
    \gG_{d}(\widehat{\mW}_d) = \frac{1}{N_d \cdot \sigma_{1}(\widehat{\mW}_d)}\sum\limits_{i=1}^{N_d^{'}}\sigma_i(\widehat{\mW}_d),
    \label{eq:stable_rank}
\end{equation}
where $\sigma_1 \geq \sigma_2 \geq \hdots \geq \sigma_{N_d}$ are low-rank singular values in descending order. Here $N_d = \rank\{\widehat{\mW}_d\}$ and the unfolding is done either on input or output channels i.e. $d\in \{3, 4\}$. We can further parameterize the stable rank $\gG_{d}$ by the HP set $\HyperParameters$, epoch $t$, and network layer $\ell$ as $\bar{\gG}_{d,t,\ell}(\HyperParameters)$. Note that $\bar{\gG}_{d,t,\ell}(\HyperParameters) \in [0, 1]$ and is used to probe CNN layers to monitor how well information is carried from input to output maps. A perfect network and set of HPs would yield $\bar{\gG}_{d, T, \ell}(\HyperParameters) = 1\quad \forall \ell \in [L]$, where $L$ is the network' number of layers and $T$ is the last epoch. In this case, each layer is a near-perfect autoencoder and the information propagation through the network is maximized. Conversely, $\bar{\gG}_{d, T, \ell}(\HyperParameters) = 0$ indicates that the information flow is very weak meaning the mapping is effectively random ($\parallel\!\mE_d\!\parallel$ is maximized). See Appendix-A for further explanation.

\subsection{Definition of New Response Function}
If $\bar{\gG}_{d, t, \ell}(\HyperParameters) = 0$ in early stages of training, no learning has occurred. This is due to the perturbing noise $\mE_d$ being fully populated and the low-rank structure $\widehat{\mW}_d$ being effectively empty. In practice, too small of an initial learning rate would also result in such a behaviour as insufficient progress has been made to reduce the randomization. We argue this becomes useful to track the number of layers with zero-valued $\gG_{d}$ as we will subsequently aim to minimize this measure across the network. This effectively becomes a measure of channel rank, and we denote this rank per epoch as
\begin{align*}
\gZ_t(\HyperParameters) \leftarrow \frac{1}{2L} \sum\limits_{\ell}^L \sum\limits_{d\in\{3,4\}}\mathbb{I}[\bar{\gG}_{d, t, \ell}(\HyperParameters) = 0]~~~
\text{where}~
\mathbb{I}[\bar{\gG}_{d, t, \ell}(\HyperParameters) = 0] = \left\{\begin{array}{ll} 1 & \textrm{if } \bar{\gG}_{d, t, \ell}(\HyperParameters) = 0 \\ 0 & \textrm{otherwise}\end{array}\right.,
\end{align*}
where $\ZPerc_t(\HyperParameters)\in [0, 1)$. We pay no attention to which layers in particular have zero-valued $\gG_{d}$; this could perhaps be explored in future work. The intuition behind averaging across all layers is to ensure that our model is ``globally'' optimized and not just locally within certain layers. Finally, we define the average rank across $T$ epochs -- which we call the \textit{global stable rank} -- as
\begin{align}
\gZ(\HyperParameters) \leftarrow \frac1T \sum\limits_{t\in[T]}\gZ_t(\HyperParameters) ~~;~\gZ(\HyperParameters)\in [0, 1).
\end{align}
This measure is therefore akin to a normalized summation of the zero-valued singular values from low-rank measures across all layers' {\color{changed}input and output} unfolded-tensor arrays over $T$ epochs. 

To solve for the optimal HP set $\HyperParameters$, we state that this is achieved when the rate of change of $\gZ(\HyperParameters)$ first goes to zero. Visually, looking at Figure \ref{aux} (and Figures C.2 \& C.3 in Appendix-C), the optimal HP is at the inception of the plateau in $\gZ(\HyperParameters)$. In support of this, we note that \citet{wilson2017marginal} found a more optimal learning rate for Adam applied to CIFAR10 to be $\num{3e-4}$ instead of the author suggested $\num{1e-3}$, and highlight how this value sits at the location we discuss here; the inception of the plateau in the rate of change of $\gZ(\HyperParameters)$. We formulate our response surface as
\begin{figure}[!thp]
\centerline{
\subfigure[Surrogate Behaviour]{\includegraphics[height=0.2\textwidth]{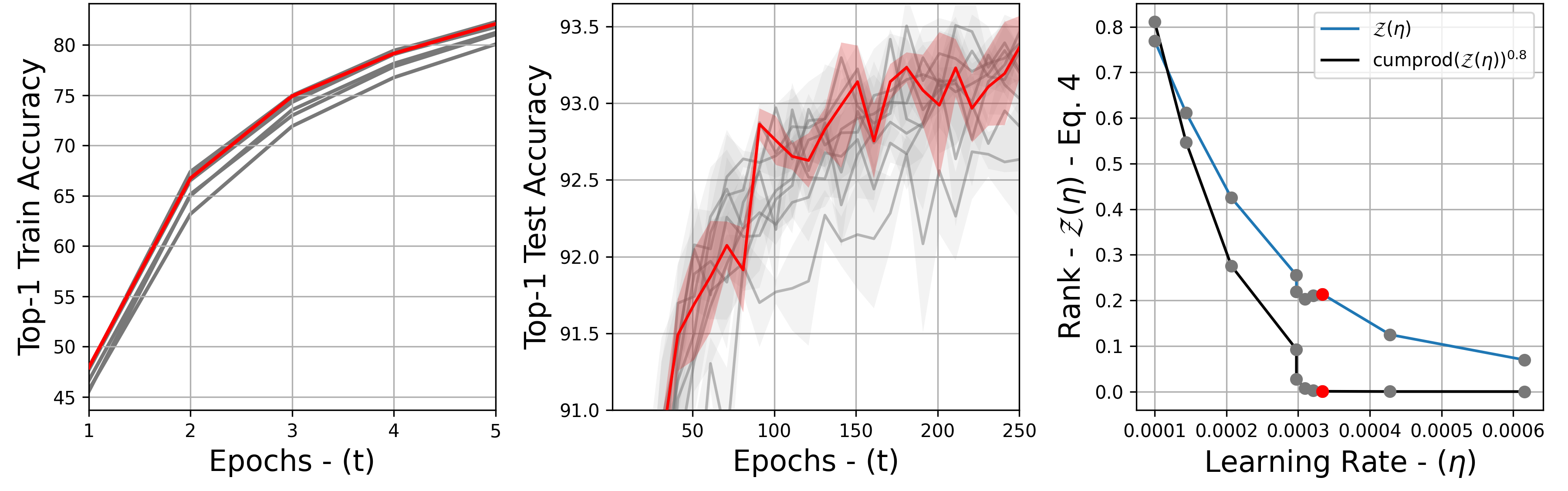}\label{aux}}
\subfigure[2D Response Surface]{\includegraphics[height=0.2\textwidth]{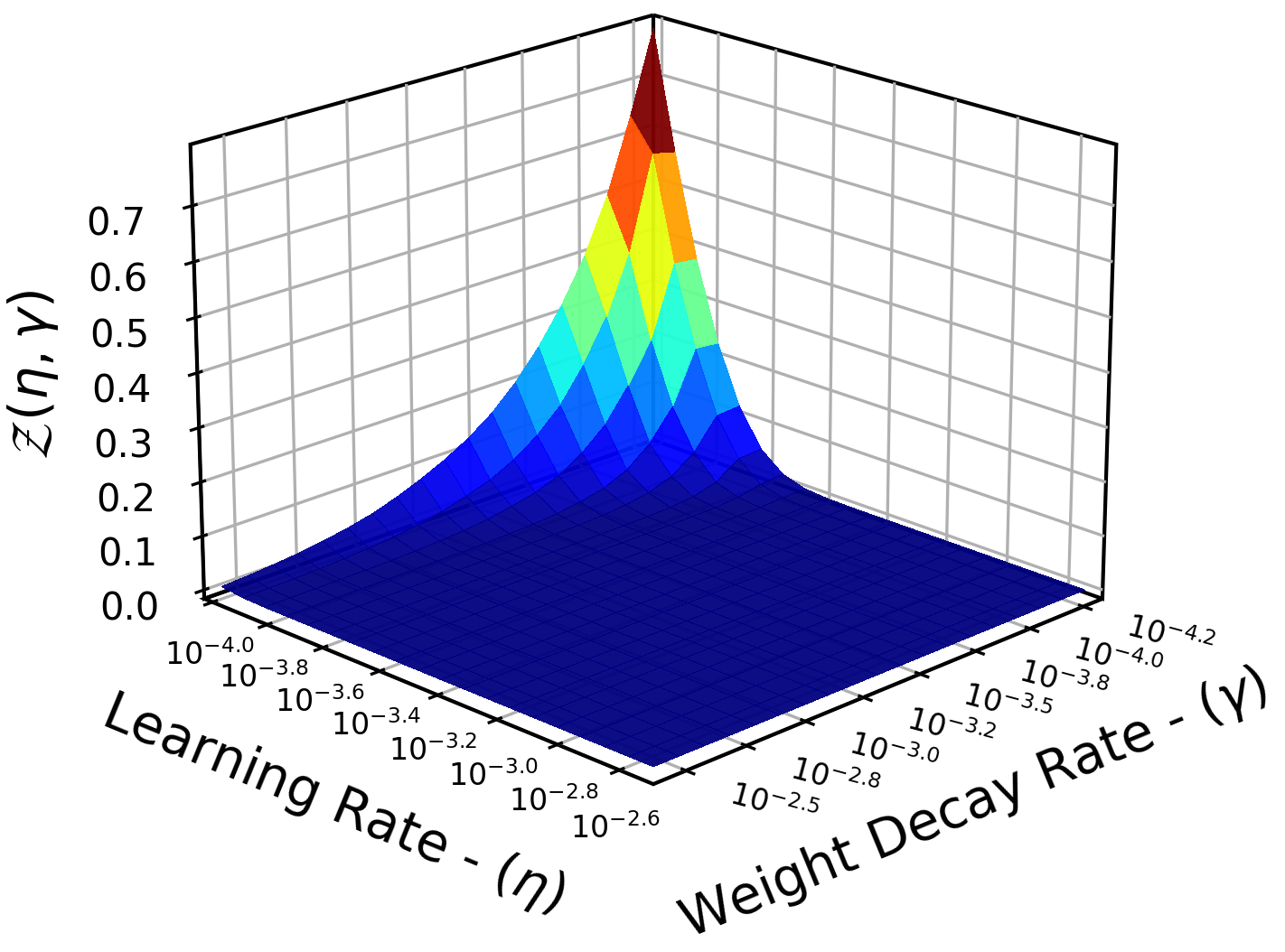} \label{figure_2d_response}}
}
\caption{(a) Surrogate behaviour of $\gZ(\GlobalLR)$ to training loss and accuracy. Red indicates the learning rate and model chosen by our method, with other trialled points in grayscale. (b) Two-dimensional response surface for $\HyperParameters = \{\GlobalLR, \gamma\}$ for ResNet34 applied to CIFAR10, trained using Adam.}
\vspace{-.2cm}
\end{figure}
\begin{align*}
    \HyperParameters^* \leftarrow \argmin_{\HyperParameters}~[\tau(\HyperParameters) = 1 - \gZ(\HyperParameters) ~\text{subject to}~ \parallel \nabla_{\HyperParameters}\gZ(\HyperParameters)\parallel_2^2 \leq \epsilon]
\end{align*}
\noindent where $\epsilon \in [0, 1)$ is a small error. We do not explicitly calculate the gradient $\nabla_{\HyperParameters}\gZ(\HyperParameters)$, but use the rate of change to guide convergence towards the inception of the plateauing region (see \autoref{sec:algorithm}). 

\section{\AlgName: Automatic HP Tuning}\label{sec:algorithm}
\begin{wrapfigure}{R}{0.5\textwidth}
  \vspace{-8mm}
    \begin{minipage}{0.5\textwidth}
\begin{algorithm}[H]
  \caption{\AlgName}
  \begin{algorithmic}[1]
  \REQUIRE{number of epochs $T=5$, starting $\HyperParameters^{(0)}$}
    \STATE $RH = [~]$ ; (empty list of rank histories)
    \STATE $j = 0$
    \WHILE{True}
        \STATE - compute trust-region $TR^{\HyperParameters^{(j)}}$ for $\HyperParameters^{(j)}$.
        \FOR{each permutation $\HyperParameters^{(j*)}$ in $TR^{\HyperParameters^{(j)}}$}
        \STATE - train for $T$ epochs, record $\ZPerc(\HyperParameters^{(j*)})$
        \ENDFOR
        \IF{each of $\gZ(\HyperParameters^{(j*)}) < 0.9$}
            \STATE - step in direction of max $\gZ(\HyperParameters^{(j*)})$
        \ELSE
            \STATE - $k\leftarrow \text{argmin}_{j*}~~\text{list of}~[\gZ(\HyperParameters^{(j*)})]$
            \STATE - $\HyperParameters^{(j+1)} \leftarrow TR^{\HyperParameters^{(j)}}[k]$
            \STATE - append smallest $\gZ(\HyperParameters^{(j*)})$ to $RH$
            \STATE - compute  $cumprod(RH)^{0.8}$
        \ENDIF
        \IF{rate of change plateaus} 
            \STATE break
        \ENDIF
        \STATE $j \mathrel{+}= 1$
    \ENDWHILE
  \end{algorithmic}
  \label{alg_algorithm}
\end{algorithm}
    \end{minipage}
    \vspace{-3mm}
\end{wrapfigure}
\textbf{How it works.} The pseudo-code for \AlgName\ is presented in Algorithm \ref{alg_algorithm}. To find the inception of the plateuing region, \AlgName\ runs a trust-region optimization algorithm, where the trust-region is formed around the HP set, and stepping is made relative to our metric, $\gZ(\HyperParameters)$. That is, at each step, \AlgName\ computes the trust-region around its current HP set $\HyperParameters$: For each HP in the set \HyperParameters, \AlgName\ scales it up and down and computes the combinatorial permutation of $\{\lambda_i / \alpha_i, \lambda_i, \lambda_i * \alpha_i~~ \forall \lambda_i \in \HyperParameters\}$, where $\alpha_i$ is a stepping constant and is set per HP. This provides a set of unique HP configurations that we denote as $TR^{\HyperParameters}$. For each configuration in this set, \AlgName\ trains for $T=5$ epochs and computes our metric $\gZ(\HyperParameters)$. Note that with caching of past values we do not need to search over each of these configurations. This trust-region optimization algorithm first steps in the direction such that $\gZ(\HyperParameters^{start}) \geq 0.9$. A step simply involves updating our HPs to the permutation that meets our given criteria, in this case $\gZ(\HyperParameters^{start}) \geq 0.9$. This start point matters since we take the cumulative product of stable ranks over trials. After, the algorithm steps in the direction to minimize $\gZ(\HyperParameters)$. At each step, this minimum $\gZ(\HyperParameters)$ is recorded. This continues until the cumulative product of the list of stable ranks $[\gZ(\HyperParameters^{start}), \hdots, \gZ(\HyperParameters^{j})]$ plateaus, where $j$ is the step count in the trust-region search.

\textbf{Choosing the trust-region size.} The choice of trust-region size will have a significant effect on the results. It should be selected be such that sequential increments of each HP should be sufficiently small, so as to not take too large a step. For \AlgName, we search around the current HPs by scaling the current HPs (up and down) by a factor of $1.5$, which is derived from similar scaling factors when doing a logarithmic grid search. This factor should be tuned for different HPs.

\textbf{Stablizing $\bm{\gZ(\HyperParameters)}.$} We calculate the rate of change of $\ZPerc(\HyperParameters)$ using the cumulative product of the sequence $[\ZPerc(\HyperParameters^0), \hdots, \gZ(\HyperParameters^j)]$, to the power of $0.8$. Since our response surface is not guaranteed to monotonically decrease, we employ the cumulative product of $[\ZPerc(\HyperParameters^0), \hdots, \gZ(\HyperParameters^j)]$, which does monotonically decrease -- since $\mathcal{Z}(\HyperParameters) \in [0, 1)$ -- to guarantee convergence. The cumulative product (to the power of $0.8$) is a good choice because (a) it dampens noise well and (b) regulates the rapid decay of the cumulative product. This power is technically tune-able, however we fix it and show in our experiments that it generalizes well. Figure C.3 in Appendix-C demonstrates further insight.

\begin{figure*}[htp]
\begin{center}
\vspace{-6mm}
    \includegraphics[width=\textwidth]{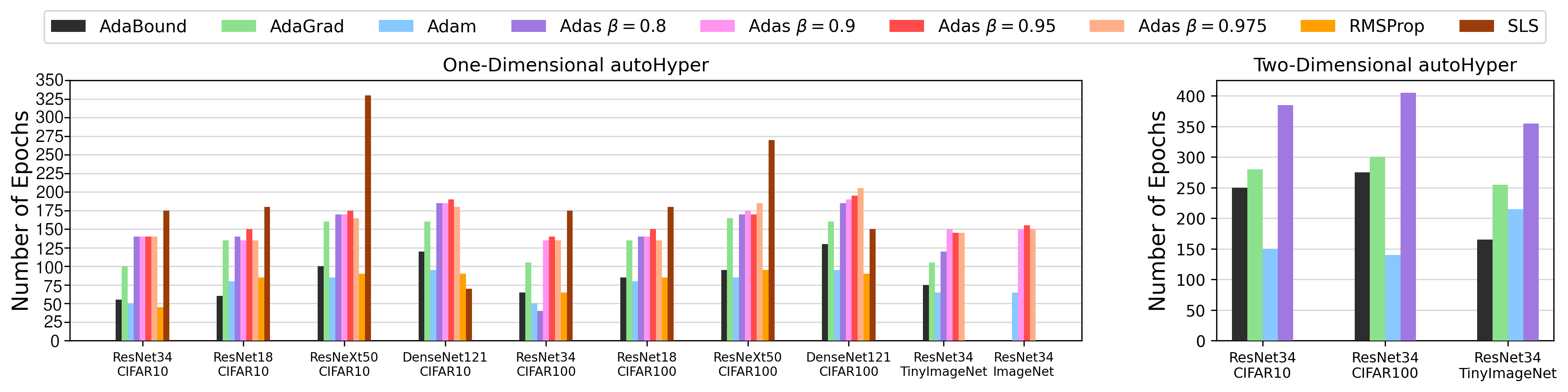}\label{trials}
\end{center}\vspace{-0.8cm}
\caption{Required number of epochs for \AlgName\ to converge over various setups.}
\label{figure_trials}
\vspace{-.5cm}
\end{figure*}

\textbf{Computational requirements.} $\gZ(\HyperParameters)$ is computed (i.e. $1$ step) using $T=5$ epochs due to stabilization of our metric after $5$ epochs. Figure \ref{figure_trials} visualizes epoch consumption for our experiments.

\section{Experiments}
\label{experiments}

\subsection{Experimental Setups}
\label{exp_setups}
\textbf{One-dimensional comparison.} We restrict our focus to the initial learning rate ($\HyperParameters = \GlobalLR$) and run experiments on CIFAR10 \citep{krizhevsky2009learning}, CIFAR100 \citep{krizhevsky2009learning}, TinyImageNet \citep{tinyimagenet}, and ImageNet \citep{imagenet2015}. On CIFAR10 and CIFAR100, we apply ResNet18 and ResNet34 \citep{he2015deep}, ResNeXt50 \citep{xie2016aggregated}, and DenseNet121 \citep{huang2017densely}. On TinyImageNet and ImageNet, we apply ResNet34. For architectures applied to CIFAR10 and CIFAR100, we train using Adam \citep{kingma2014adam}, AdaBound \citep{luo2019adaptive}, Adas\textsuperscript{($\beta = \{0.8, 0.9, 0.95, 0.975\}$)} \citep{hosseini2020adas} (with early-stop), AdaGrad \citep{duchi2011adaptive}, RMSProp \citep{tieleman2012lecture}, and SLS \citep{vaswani2019sls}. For ResNet34 applied to TinyImageNet, we train using Adam, AdaBound, AdaGrad, and Adas\textsuperscript{($\beta = \{0.8, 0.9, 0.95, 0.975\}$)}. For ResNet34 applied to ImageNet, we train using Adam and Adas\textsuperscript{($\beta = \{0.9, 0.95, 0.975\}$)}. Further, we conduct baselines using the author suggested learning rates, RS generated learning rates, and BO generated learning rates. Only Adam, AdaBound, AdaGrad, and Adas\textsuperscript{$0.9$} are used in the RS and BO baselines. Each experiment is run for $5$ randomly initialized trials, each for 250 epochs, and we report averages.

\textbf{Two-dimensional Comparison.} We restrict our focus to the initial learning rate and weight decay rate ($\HyperParameters = \{\GlobalLR, \gamma\}$) and run experiments on CIFAR10, CIFAR100, and TinyImageNet. We apply ResNet34 and train using Adam, AdaBound, AdaGrad, and Adas\textsuperscript{$\beta = 0.9$}. Our baseline is composed only of BO generated initial learning rate and weight decay rate. Each experiment is run 5 times from randomly initialized starting points, each for 250 epochs, and we report averages.

\textbf{Random Search and Bayesian Optimization setup.} Because RS and BO are highly sensitive to the manually set search spaces \citep{choi2020empirical, sivaprasad2020optimizer}, we attempt a fair comparison by providing similar search spaces that \AlgName\ is designed around. That is, for learning rate, $\GlobalLR_{min} = \num{1e-4}$ and $\GlobalLR_{max} = 0.1$ and for weight decay rate, $\gamma_{min} = \num{1e-7}$ and $\gamma_{max} = 0.1$.  Both RS and BO are given the same computational budget that \AlgName\ had for each experiment (see Figure \ref{figure_trials}). This does provide the RS and BO experiments with a slight advantage since a priori knowledge of how many epochs and trials to consider is not provided to \AlgName. Further, RS and BO are given the advantage of using the test set for evaluation, since CIFAR10/CIFAR100 do not have an explicit development set. This was also done for TinyImageNet. 

\textbf{Additional notes.} For BO, we used the Adaptive Experimentation Platform (\url{https://ax.dev/} \citep{balandat2020botorch}). Additional results and details are in Appendix-D and Appendix-E. 

\textbf{A note on multi-fidelity techniques.} We attempted experiments on HyperBand \cite{li2017hyperband} and BOHB \cite{falkner2018bohb} using the HPBandSter library but found that tuning the number of iterations (e.g. for successive halving) and the allocated budget to be difficult and iterative, and performance was significantly worse. For fairness, those experiments were not completed, as they required much more tuning.

\subsection{Results}\label{sec_results}
\subsubsection{One-Dimensional Comparison}
\textbf{Consistency across experimental setup.} Table \ref{table_results} tells us that our method generalizes well to experimental setups. If there is loss of performance when using an initial learning rate generated by \AlgName, this loss is $<1\%$ in all experiments except three: On CIFAR100, the author baselines of ResNeXt50 trained using Adam, ResNext50 trained using RMSProp, and DenseNet121 trained using AdaBound achieve $1.2\%$, $2.28\%$ and $2.3\%$ better top-1 test accuracy, respectively. 
\begin{table}[htp]
    \setlength\tabcolsep{1pt} 
	\caption{Final epoch (250) top-1 test accuracies for $\HyperParameters = \GlobalLR$. Values marked with a `*' indicate early-stopping. The best result is highlighted in {\color{best}green}, and for \AlgName\ results, {\color{close}orange} highlights when the results lie within the standard deviation from the best.}
	\label{table_results}
	\footnotesize{
		\begin{tabular}{c|ccc|c||ccc|c}
		\hlinewd{1pt}
      &\multicolumn{4}{c||}{ResNet34 on TinyImageNet}&\multicolumn{4}{c}{ResNet34 on CIFAR100}\\
    \hlinewd{1pt}
      Optimizer&Author&RS&BO&\AlgName&Author&RS&BO&\AlgName\\
		\hline 
		\hline
		AdaBound&$55.48_{\pm0.67}$&$54.88_{\pm0.57}$&$55.18_{\pm0.13}$&${\color{best}\bm{56.22_{\pm0.17}}}$&$71.94_{\pm0.66}$&$73.17_{\pm0.09}$&${\color{best}\bm{73.34_{\pm0.50}}}$&${\color{close}\bm{73.15_{\pm0.24}}}$\\
    AdaGrad&${\color{best}\bm{55.81_{\pm0.84}}}$&$50.66_{\pm0.33}$&$51.26_{\pm0.55}$&${\color{close}\bm{55.04_{\pm0.54}}}$&$67.02_{\pm0.23}$&$66.02_{\pm0.59}$&$67.11_{\pm0.54}$&${\color{best}\bm{67.43_{\pm0.59}}}$\\
		Adam&$52.13_{\pm1.14}$&$54.86_{\pm0.21}$&${\color{best}\bm{54.96_{\pm0.39}}}$&${\color{close}\bm{54.46_{\pm1.14}}}$&$71.11_{\pm0.37}$&$70.55_{\pm0.15}$&$70.94_{\pm0.62}$&${\color{best}\bm{71.43_{\pm0.28}}}$\\
		Adas\textsuperscript{$0.9$}&$59.91_{\pm0.45}$&$59.76_{\pm0.50}$&$59.52_{\pm0.19}$&${\color{best}\bm{61.56_{\pm0.58}}}$&${\color{best}\bm{75.99_{\pm0.09}}}$&$73.51_{\pm0.13}$&$69.38_{\pm0.62}$&${\color{close}\bm{75.78^*_{\pm0.21}}}$\\
        \hlinewd{1pt}
      &\multicolumn{4}{c||}{ResNet18 on CIFAR100}&\multicolumn{4}{c}{DenseNet121 on CIFAR100}\\
    \hlinewd{1pt}
      Optimizer&Author&RS&BO&\AlgName&Author&RS&BO&\AlgName\\
		\hline 
		\hline
      AdaBound&$72.04_{\pm0.30}$&${\color{best}\bm{73.24_{\pm0.30}}}$&$73.28_{\pm0.21}$&${\color{close}\bm{\bm{73.16_{\pm0.25}}}}$&$68.90_{\pm0.36}$&${\color{best}\bm{69.30_{\pm0.22}}}$&$68.91_{\pm0.23}$&$67.00_{\pm0.20}$\\
      AdaGrad&${\color{best}\bm{67.76_{\pm0.50}}}$&$67.75_{\pm0.24}$&$67.40_{\pm0.43}$&${\color{close}\bm{67.75_{\pm0.56}}}$&$62.14_{\pm0.15}$&$62.43_{\pm0.83}$&$62.57_{\pm0.65}$&${\color{best}\bm{62.79_{\pm0.35}}}$\\
      Adam&$70.34_{\pm0.27}$&$70.58_{\pm0.24}$&${\color{best}\bm{71.03_{\pm0.42}}}$&$70.09_{\pm0.35}$&$67.48_{\pm0.17}$&$67.96_{\pm0.23}$&$68.58_{\pm0.64}$&${\color{best}\bm{69.05_{\pm0.49}}}$\\
      Adas\textsuperscript{$0.9$}&$75.15_{\pm0.17}$&$75.11_{\pm0.18}$&$75.13_{\pm0.23}$&${\color{best}\bm{75.27^*_{\pm0.28}}}$&${\color{best}\bm{73.25_{\pm0.25}}}$&$73.22_{\pm0.30}$&$72.89_{\pm0.20}$&${\color{close}\bm{73.13^*_{\pm0.44}}}$\\
        \hlinewd{1pt}
		\end{tabular}
		\vspace{-2mm}
	}
\end{table}

\textbf{Improved performance over RS and BO.} We highlight how \AlgName\ is able to generalize over experimental setup whereas RS and BO cannot. In particular, RS and BO applied on AdaGrad and Adas\textsuperscript{$0.9$} struggle to compete with \AlgName, particularly in more complex datasets such as CIFAR100. This is most evident in ResNet34 applied to CIFAR100 (see Table \ref{table_results}). This highlights how \AlgName\ can automatically find more competitive learning rates to RS and BO and without any manual intervention. Most interestingly, RS and BO were given a large advantage in that testing accuracy was used in their implementation rather than the conventional validation accuracy, and yet \AlgName's performance (which uses training data) is maintained.

\begin{wraptable}{r}{0.54\textwidth}
\vspace{-2mm}
\setlength\tabcolsep{1pt}
\caption{ImageNet test accuracies for $\HyperParameters = \GlobalLR$.}
\footnotesize{
    \centering
    \begin{tabular}{c|cccc}
    \hlinewd{1pt}
      &\multicolumn{4}{|c}{ResNet34 on ImageNet}\\
    \hlinewd{1pt}
      Method&Adam&Adas\textsuperscript{$0.9$}&Adas\textsuperscript{$0.95$}&Adas\textsuperscript{$0.975$}\\
      \hline\hline
		Author&$63.75_{\pm0.08}$&${\color{best}\bm{72.11_{\pm0.16}}}$&$73.05_{\pm0.17}$&${\color{best}\bm{72.52_{\pm0.03}}}$\\
    \AlgName&${\color{best}\bm{68.68_{\pm0.24}}}$&${\color{close}\bm{71.87^*_{\pm0.20}}}$&${\color{best}\bm{73.09_{\pm0.23}}}$&${\color{close}\bm{72.42_{\pm0.15}}}$\\
    \hlinewd{1pt}
    \end{tabular}
    }
    \label{tab:my_label}
    \vspace{-3mm}
\end{wraptable}

\textbf{ImageNet/TinyImageNet Improvements.} We highlight how well \AlgName\ performs when applied to TinyImagetNet and ImageNet. ResNet34 trained using Adam and applied to TinyImageNet and ImageNet achieves final improvements of $3.14\%$ and $4.93\%$ in top-1 test accuracy, respectively, shown in Table \ref{tab:my_label}. Such improvements come at a minimal cost using our method, requiring 65 epochs (4 hours) and 80 epochs (59 hours) for TinyImageNet and ImageNet, respectively (Figure \ref{figure_trials}).

\textbf{Fast and consistent convergence rates.} We visualize the convergence rates of our method in Figure \ref{figure_trials}. Importantly, we identify \AlgName's consistency in required epochs per optimizer across architecture and dataset selection as well as the low convergence times. Our method exhibits less consistent results when optimizing using SLS as SLS tends to result in high $\gZ(\GlobalLR)$ over multiple epochs and different learning rates. Despite this, our \AlgName\ still converges and performs well.

\begin{table}[!h]
    \setlength\tabcolsep{1pt} 
	\caption{Final epoch (250) top-1 test accuracies $\HyperParameters = \{\GlobalLR, \gamma\}$.}
	\label{table_results_2d}
	\vspace{-5mm}
	\begin{center}
	\footnotesize{
		\begin{tabular}{c|cccc||cccc}
		\hlinewd{1pt}
		&\multicolumn{4}{c||}{ResNet34 on TinyImageNet}&\multicolumn{4}{c}{ResNet34 on CIFAR10}\\
		\hlinewd{1pt}
		Method&AdaBound&AdaGrad&Adam&Adas\textsuperscript{$0.9$}&AdaBound&AdaGrad&Adam&Adas\textsuperscript{$0.9$}\\
		\hline 
		\hline
        BO&$54.92_{\pm0.30}$&$49.56_{\pm0.54}$&$53.45_{\pm0.56}$&$58.17_{\pm0.32}$&${\color{best}\bf{93.21_{\pm0.16}}}$&$90.50_{\pm0.16}$&$93.18_{\pm0.25}$&$91.52_{\pm0.08}$\\
        \hline
        \AlgName&${\color{best}\bf{57.02_{\pm0.20}}}$&${\color{best}\bf{55.59_{\pm0.71}}}$&${\color{best}\bf{55.29_{\pm0.20}}}$&${\color{best}\bf{58.19_{\pm0.20}}}$&${\color{close}\bf{92.84_{\pm0.28}}}$&${\color{best}\bf{91.49_{\pm0.21}}}$&${\color{best}\bf{93.24_{\pm0.06}}}$&${\color{best}\bf{92.64_{\pm0.18}}}$\\
		\hlinewd{1pt}
		\end{tabular}
		}
		\end{center}
\vspace{-7mm}
\end{table}
\subsubsection{Two-Dimensional Comparison}
\textbf{Improved performance over Bayesian Optimization.} Analyzing Table \ref{table_results_2d}, we see that \AlgName\ outperforms BO in the two-dimensional case. In particular, there is a $3.74\%$ improvement for ResNet34 applied to CIFAR100 using Adam. Of the $12$ experiments, there are only $3$ where BO is able to outperform \AlgName. We note that, of these $3$ cases, \AlgName\ is within the standard deviation of error in all but two: ResNet34 applied to CIFAR100 using AdaGrad and Adas\textsuperscript{$0.9$}.

\textbf{Improvements in  TinyImageNet.} We highlight how \AlgName\ is able to significantly outperform BO on the more complex TinyImageNet dataset. In particular,  ResNet34 applied to TinyImageNet, \AlgName\ achieves $2.1\%$, $6.03\%$, and $1.84\%$ improvement when using AdaBound, AdaGrad, and Adam, respectively. This result is very promising as complex datasets are often the most difficult and time consuming datasets to perform HPO on.


\section{Conclusion}
In this introductory work, we explored a new class of hyper-parameter optimization and proposed an analytical response surface that acts as a surrogate to validation metrics and generalizes well. We proposed an algorithm, \AlgName, that optimizes for this surface and progresses towards fully automatic multi-dimensional HPO. \AlgName\ is able to, on average, outperform existing SOTA and only requires training data. In future works, we would like to expand beyond the two-dimensional case and explore further developments to our metric. Further, we would like to research ways in eliminating the current internal hyper-parameters as well improvements in computational complexities, particularly when applied to the multi-dimensional case.

\bibliography{citations}
\pagebreak
\appendix
\renewcommand\thefigure{\thesection.\arabic{figure}}    
\setcounter{figure}{0}
\section{Stable Rank Optimality}
\label{sec:app_explained}
Given the definitions of stable rank in \autoref{eq:stable_rank} , we argue that a stable rank of $1$ indicates a perfectly learned network. Specifically, higher values $\gG(\widehat{\tW}_d) \rightarrow 1$ indicates that most singular values are non-zero (i.e. $\sigma_i^2(\widehat{\tW}_d) > 0 \forall i \in [1, \hdots, n']$ where $n' \rightarrow n$. This creates a subspace spanned by a set of independent vectors corresponding to the non-zero singular values mentioned above. In other words, $\gG(\widehat{\tW}_d) \rightarrow 1$ corresponds to a many-to-many mapping but not a many-to-low (i.e. rank-deficient) mapping. Also, note that the stable rank is measured on the low-rank and not the raw measure of the weights. So the higher value indicates that the \textit{learned} weight matrix contains more non-empty structure which can be interpreted as a sign of a meaningful learning.

\section{Rank behaviour over multiple epochs}
\label{app_rank}
Here we present the behaviour of $\gZ_t(\GlobalLR)$.
\begin{figure}[h]
\centering
    \subfigure[Adam]{\includegraphics[height=0.15\textwidth]{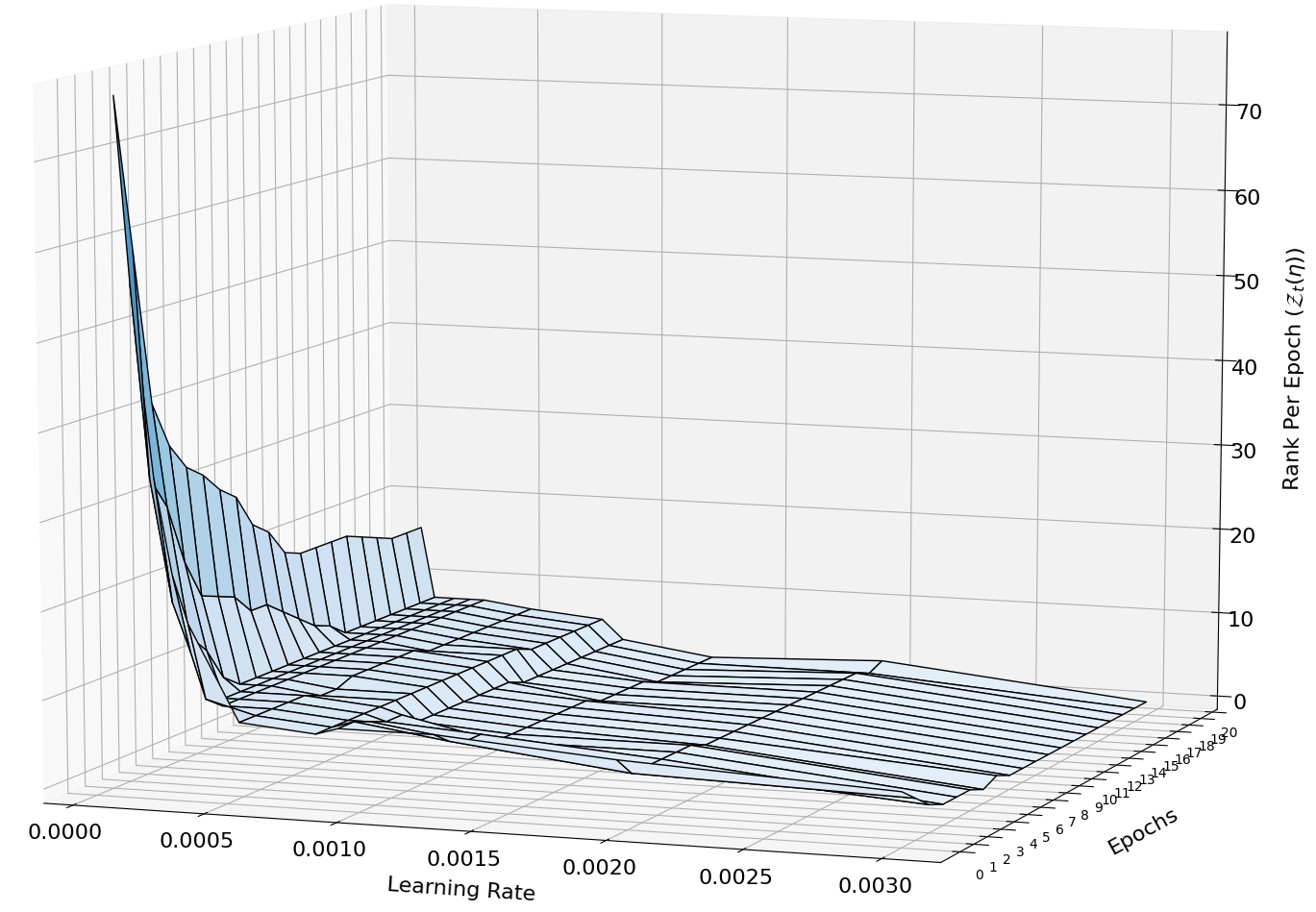}}
    \subfigure[AdaGrad]{\includegraphics[height=0.15\textwidth]{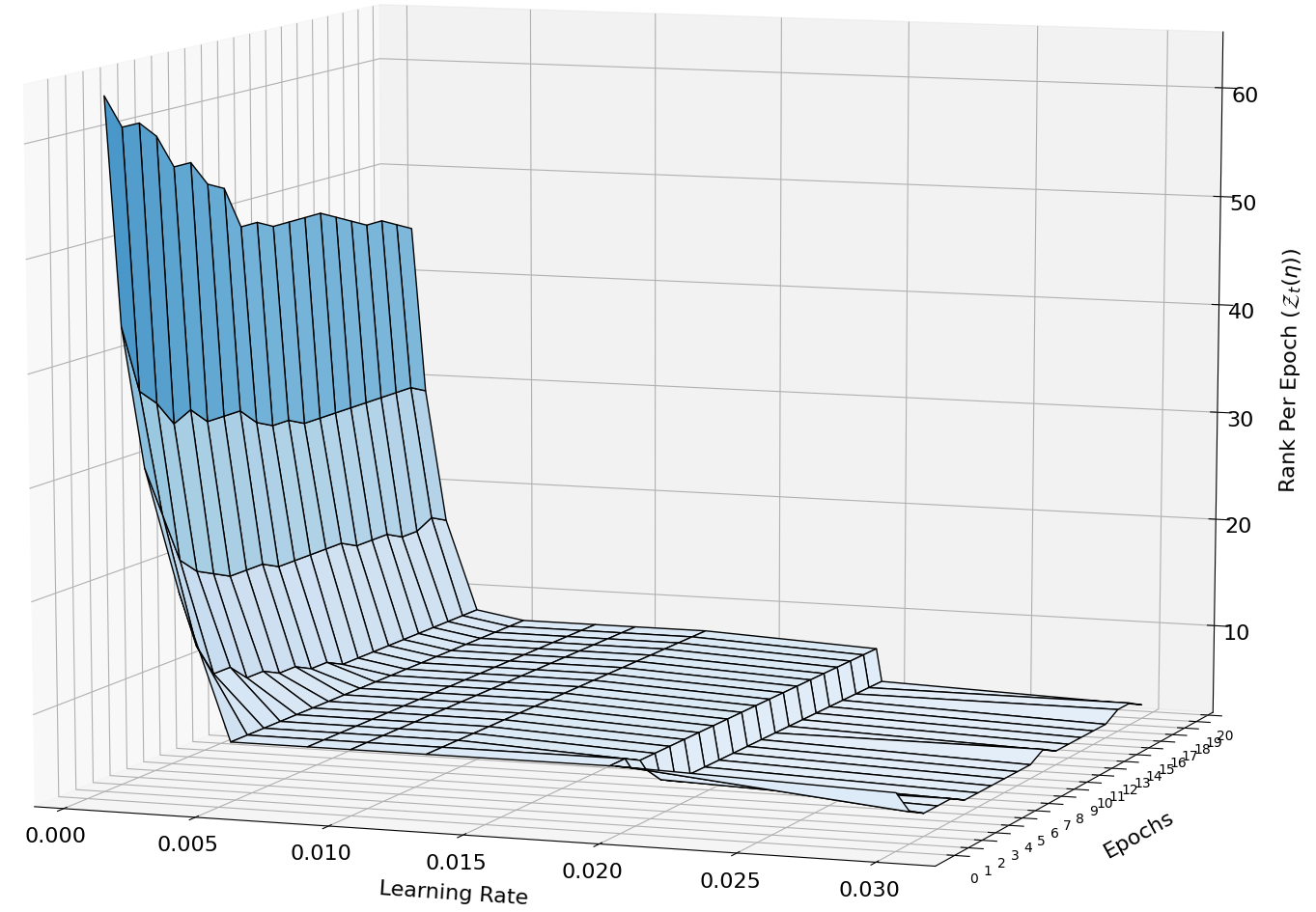}}
    \subfigure[Adas $\beta=0.8$]{\includegraphics[height=0.15\textwidth]{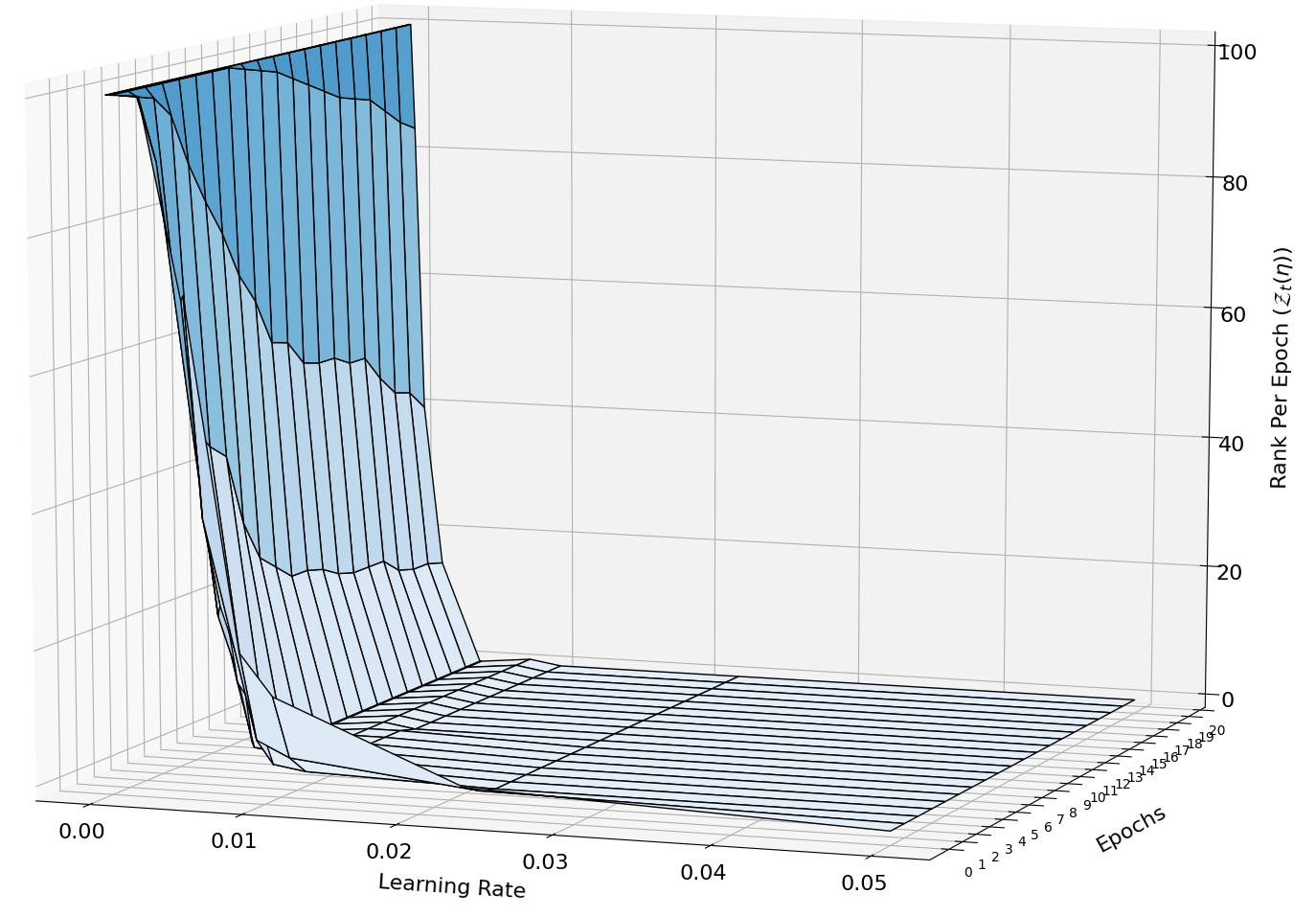}}
    \subfigure[RMSProp]{\includegraphics[height=0.15\textwidth]{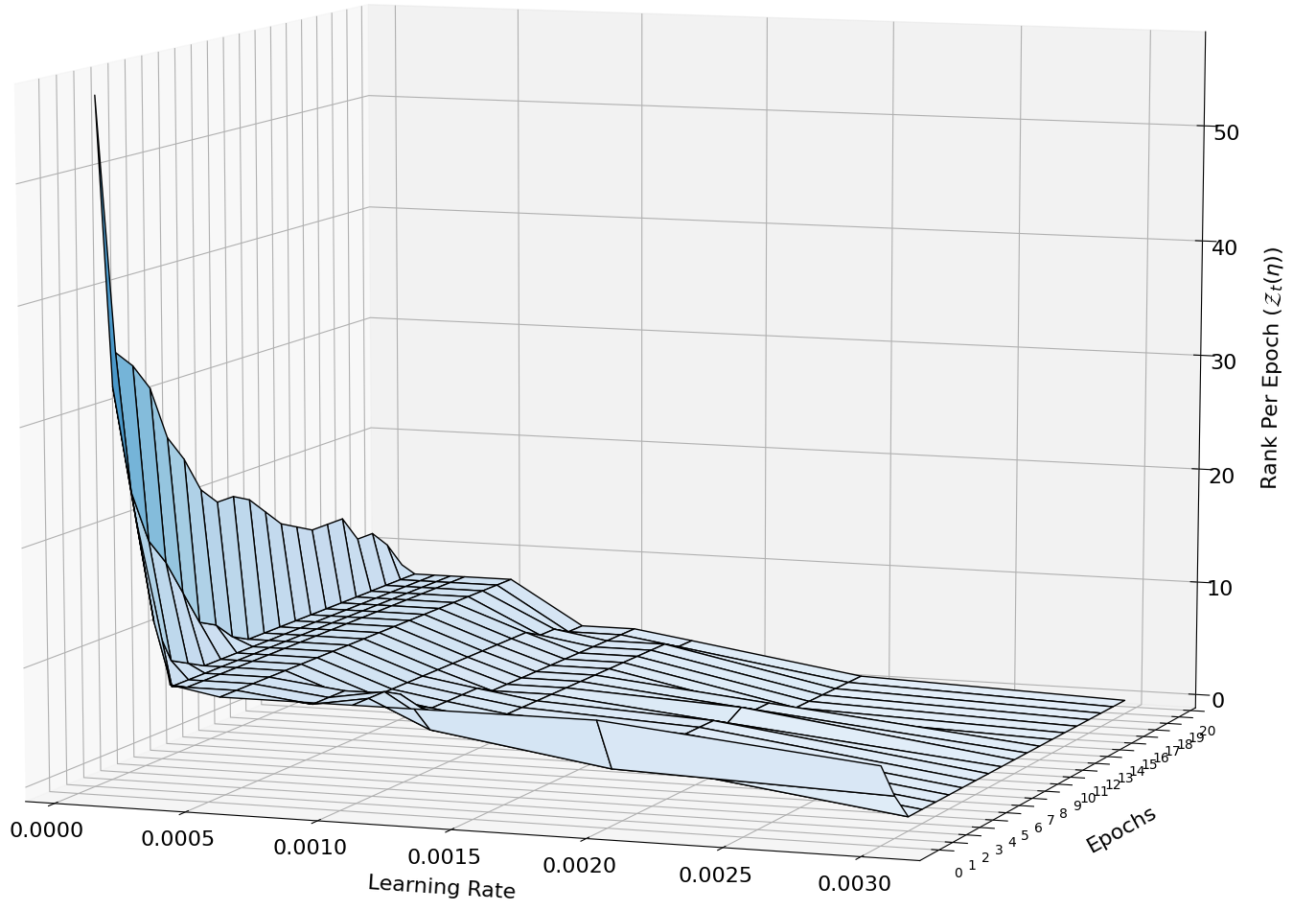}}\\
    \subfigure[Adam]{\includegraphics[height=0.15\textwidth]{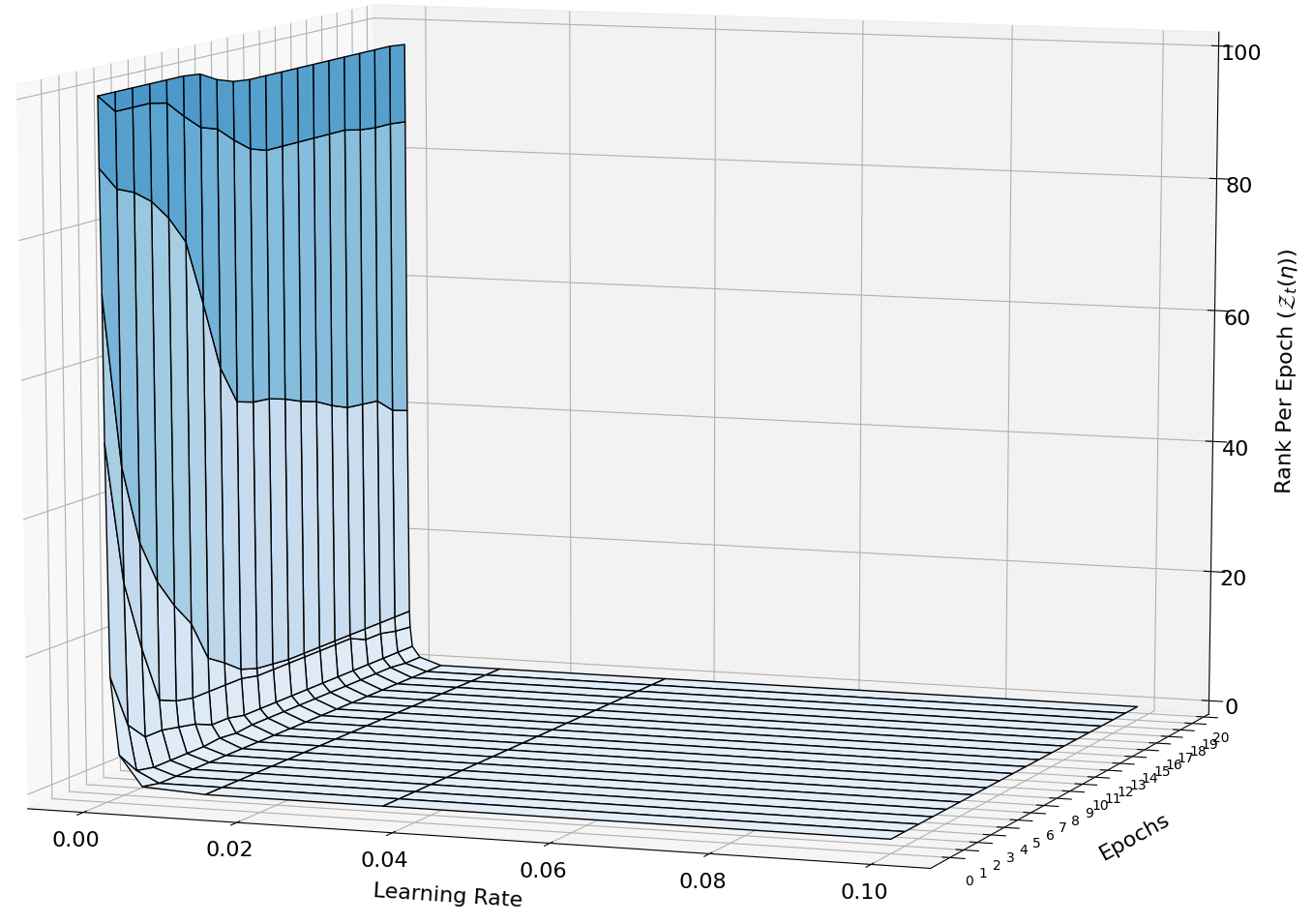}}
    \subfigure[AdaGrad]{\includegraphics[height=0.15\textwidth]{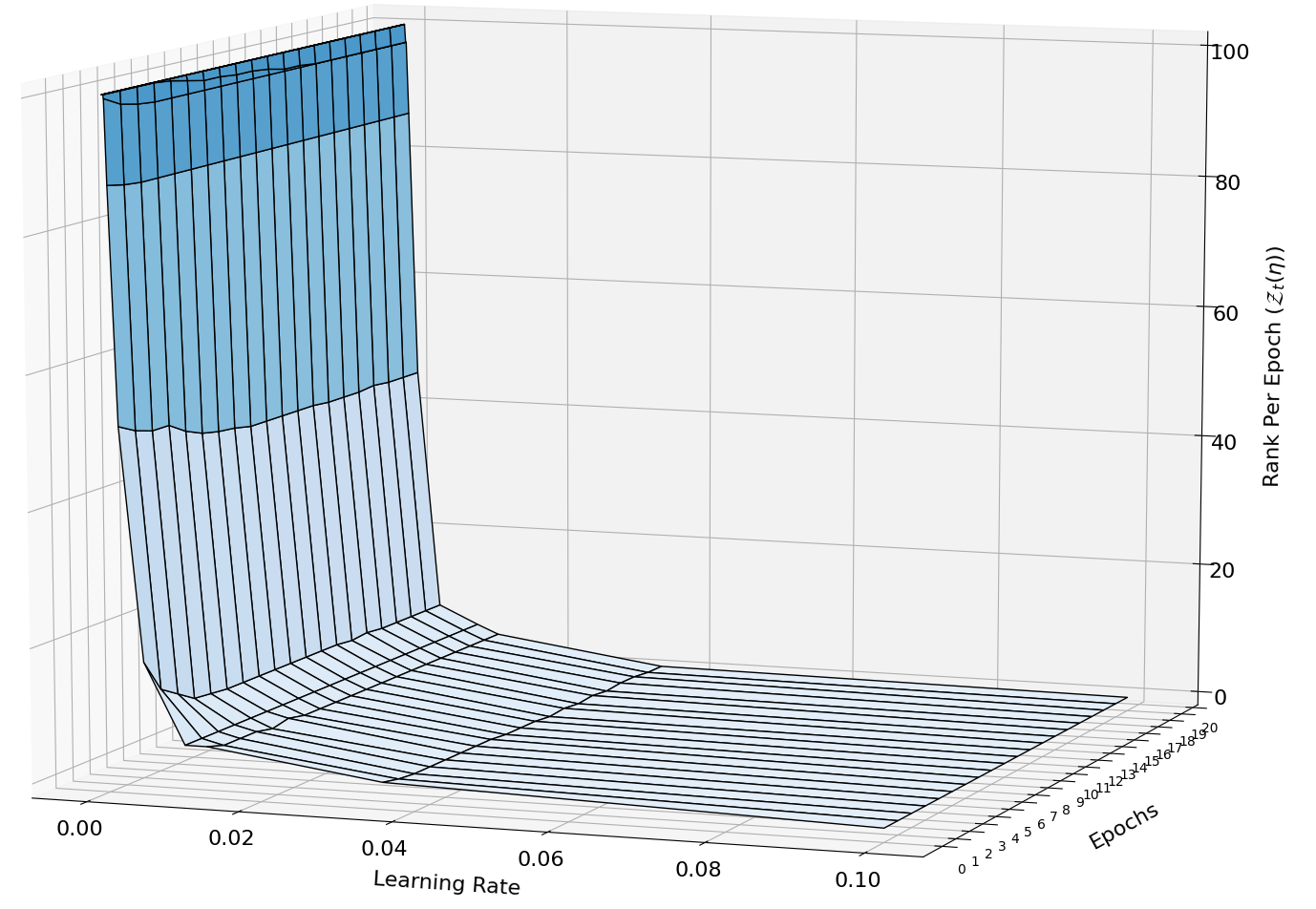}}
    \subfigure[Adas $\beta=0.8$]{\includegraphics[height=0.15\textwidth]{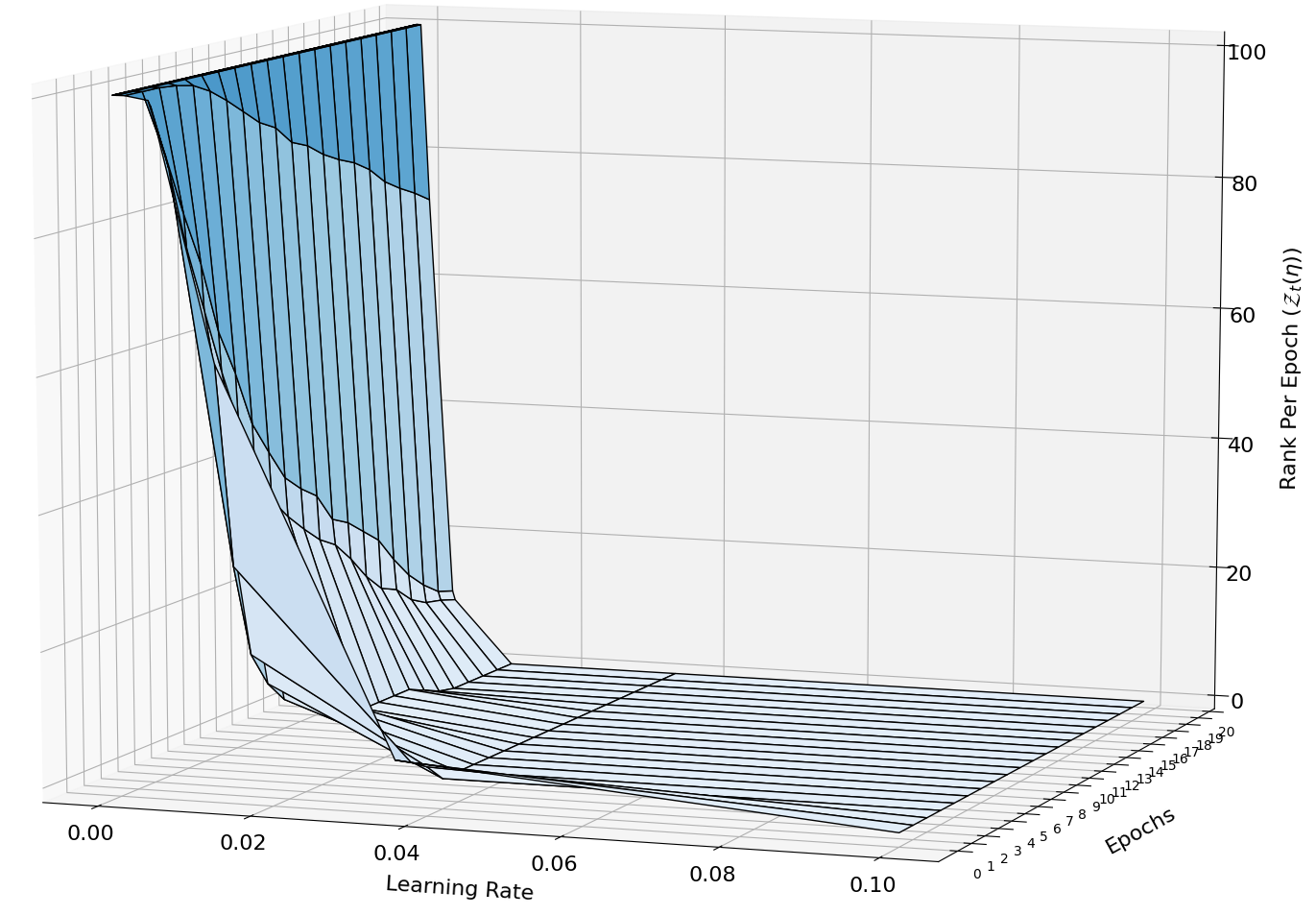}}
    \subfigure[RMSProp]{\includegraphics[height=0.15\textwidth]{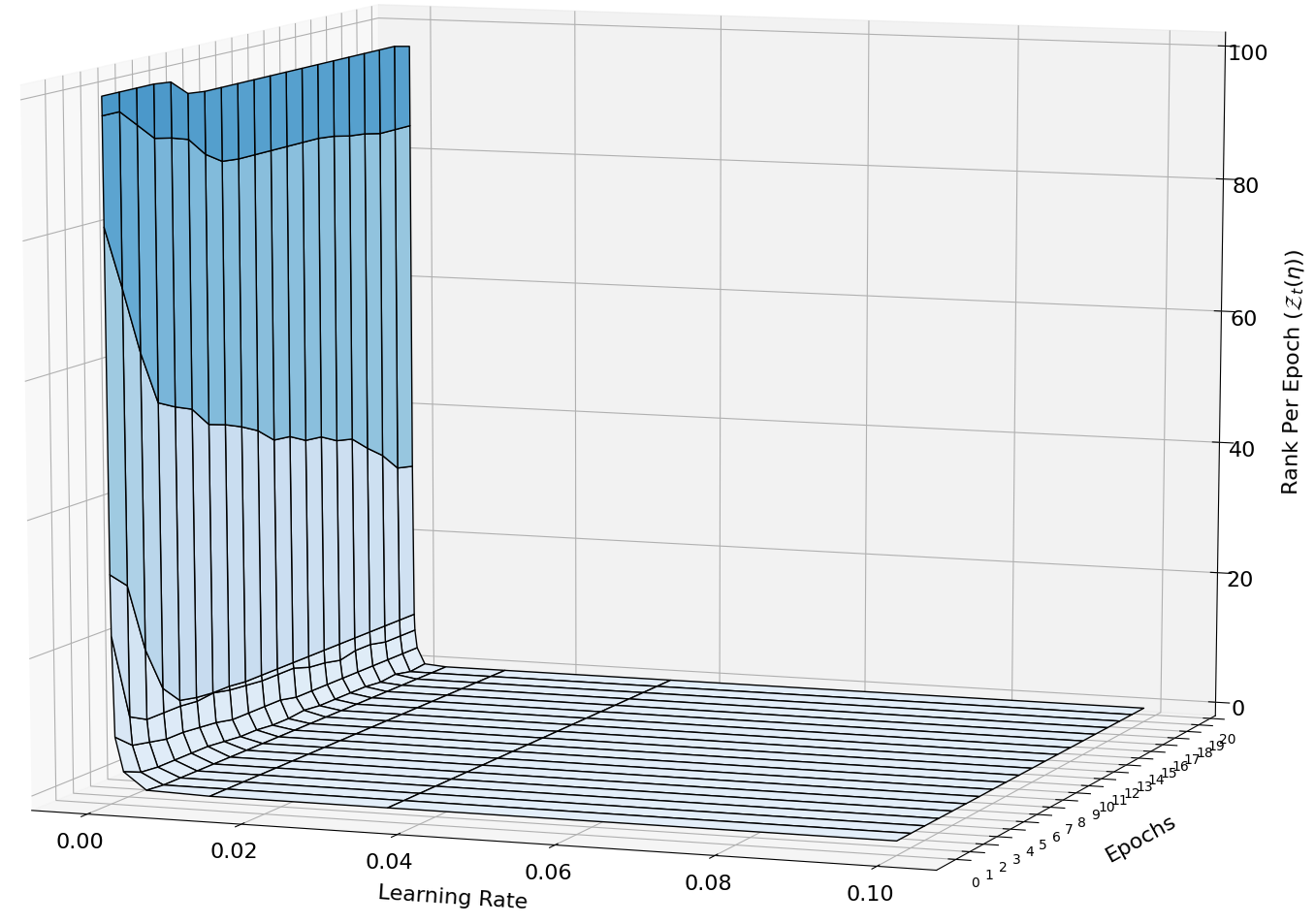}}
\caption{Rank ($\gZ_t(\GlobalLR)$) for various learning rates on VGG16 trained using Adam, AdaGrad, Adas $\beta=0.8$, and RMSProp and applied to CIFAR10. A fixed epoch budget of 20 was used. We highlight how across these 20 epochs, very little progress is made beyond  the first first epochs. It is from this analysis that we choose our epoch range of $T=5$.}
\label{figure_justification_vgg16}
\end{figure}
\section{Additional Figures for Response Surface}
\begin{figure}[t]
\centerline{
\includegraphics[height=0.22\textwidth]{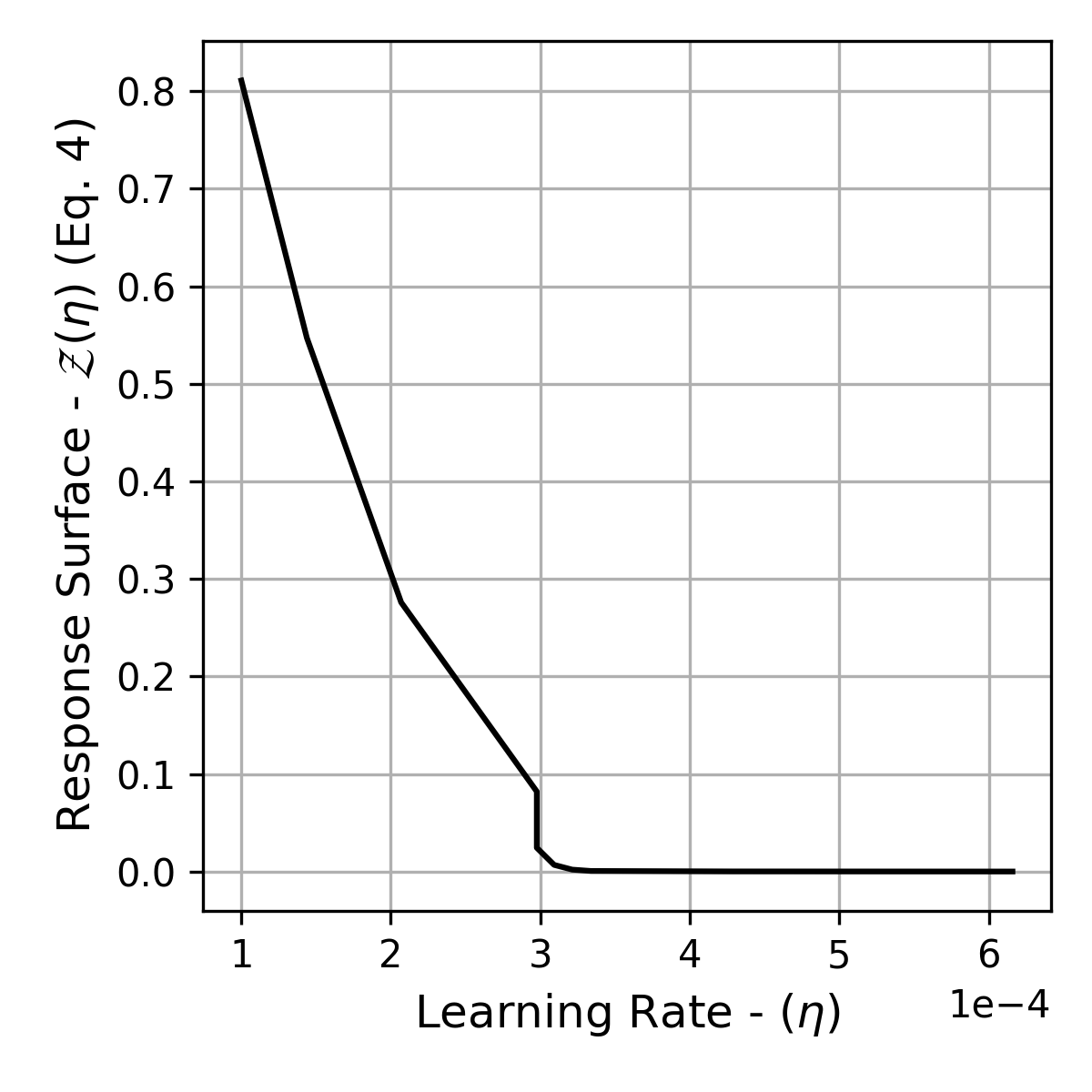}
\includegraphics[height=0.22\textwidth]{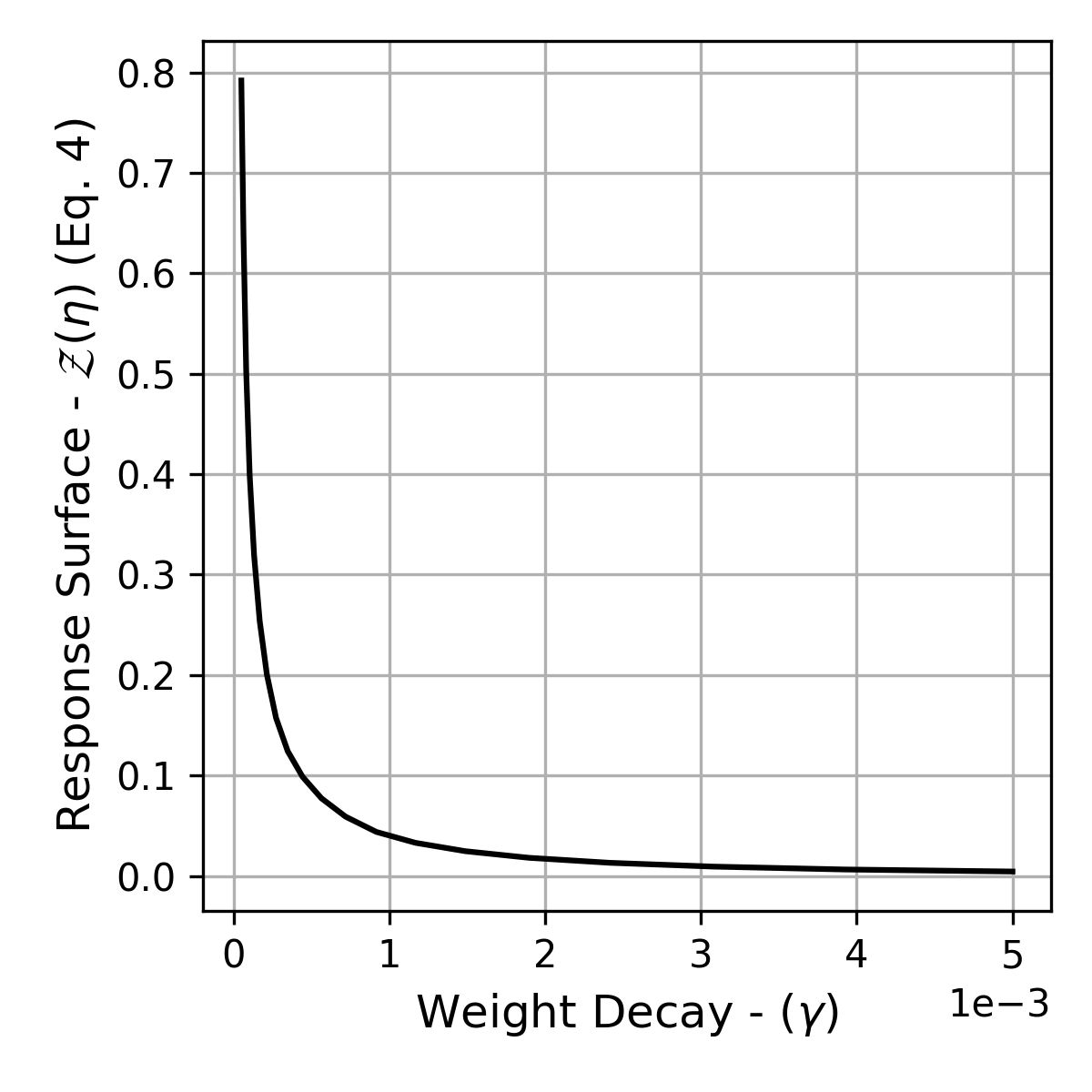}
    }
\caption{Behaviour of our metric $\gZ(\HyperParameters)$ in response to initial learning rate and weight decay. Note that we plot the regularized cumulative product of $\gZ(\HyperParameters)$ here.}
\vspace{-.5cm}
\label{response_lr_wd}
\end{figure}

\begin{figure}[h]
\centering
    \includegraphics[height=0.22\textwidth]{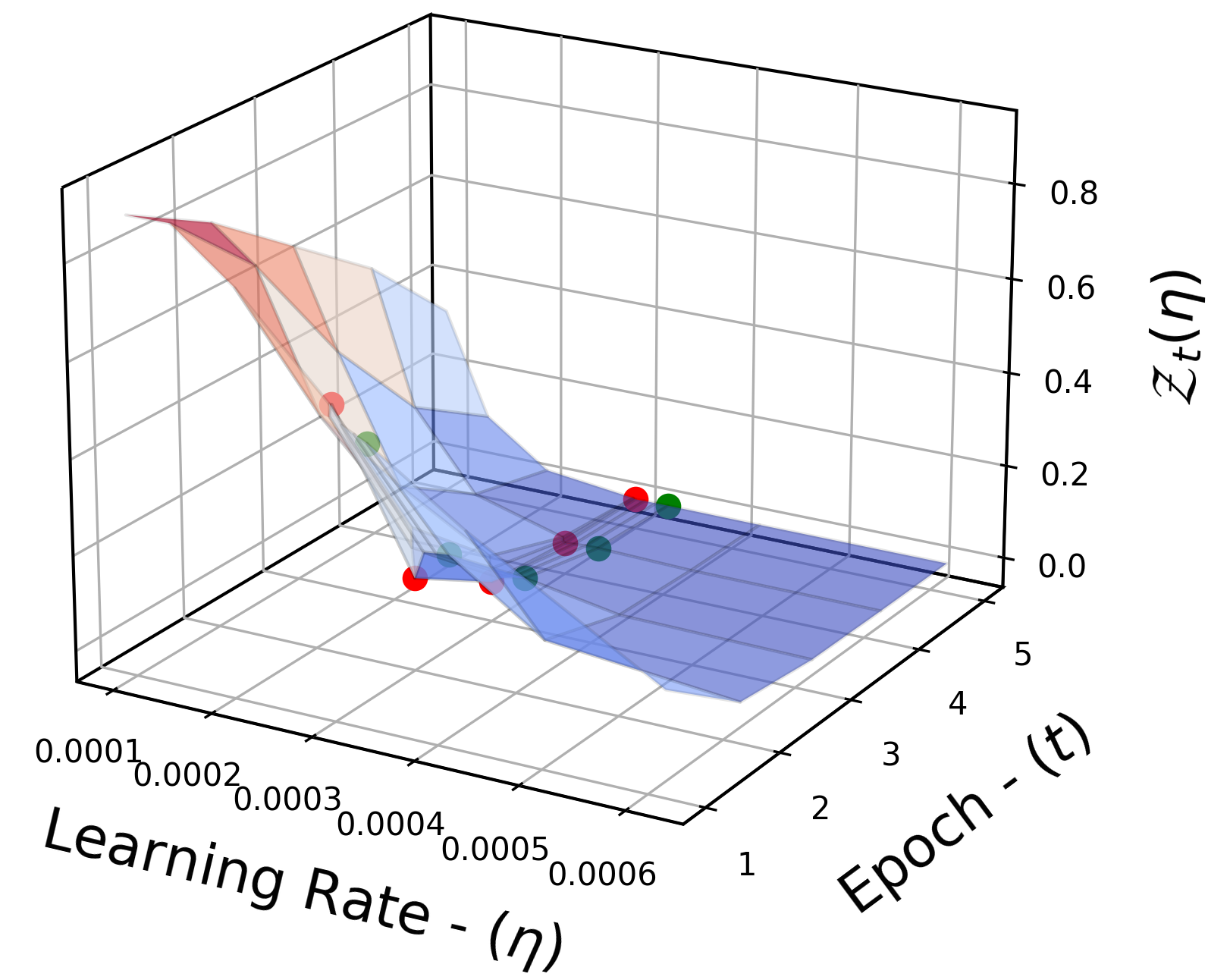}
\caption{$\gZ_t(\GlobalLR)$ for various learning rates using Adam on ResNet34 applied to CIFAR10. The author-suggested initial learning rate is indicated by the red markers, and the \AlgName\ suggested learning rate is indicated by the green markers.}
\label{figure_justification}
\vspace{-.1cm}
\end{figure}
\begin{figure}[h]
\centering
    \subfigure[ResNet34/Adam]{\includegraphics[width=.22\textwidth]{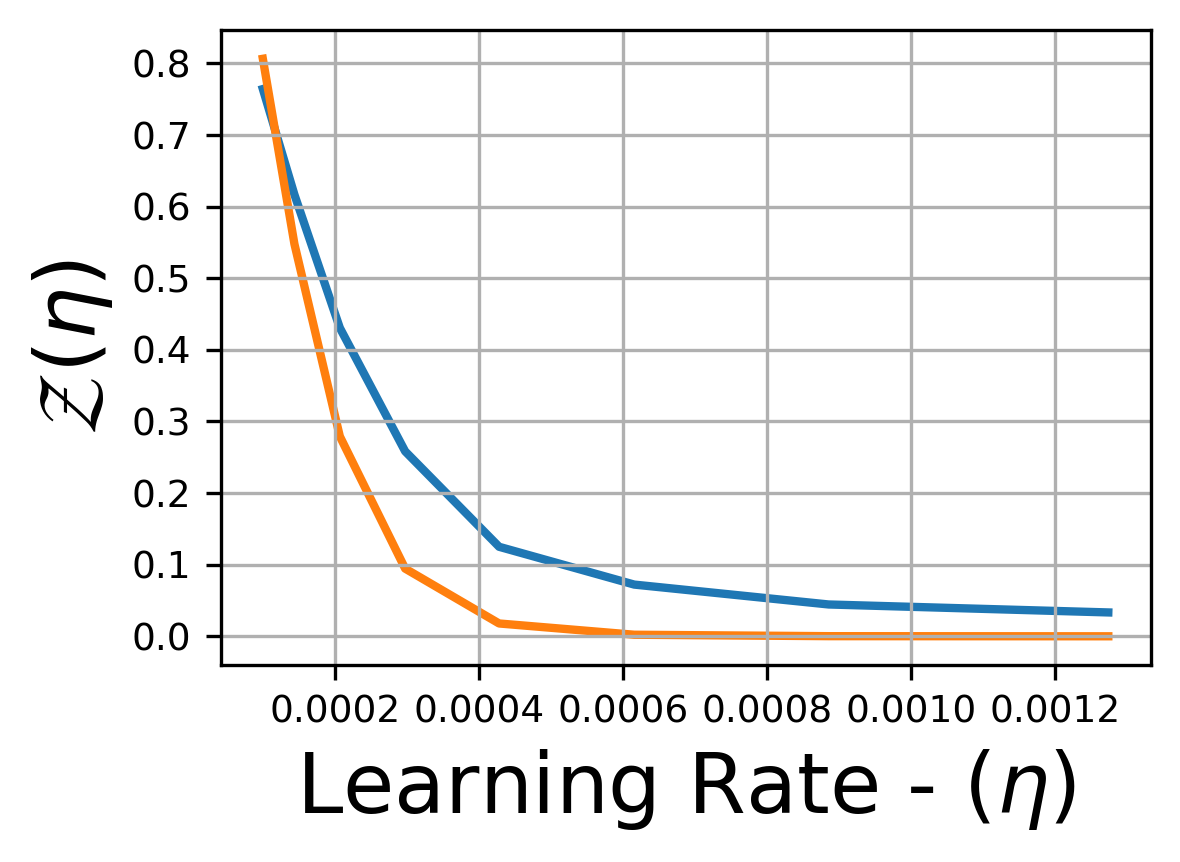}\label{figure_stable}}
    \subfigure[EffNetB0/Adas\textsuperscript{$\beta=0.8$}]{\includegraphics[width=0.22\textwidth]{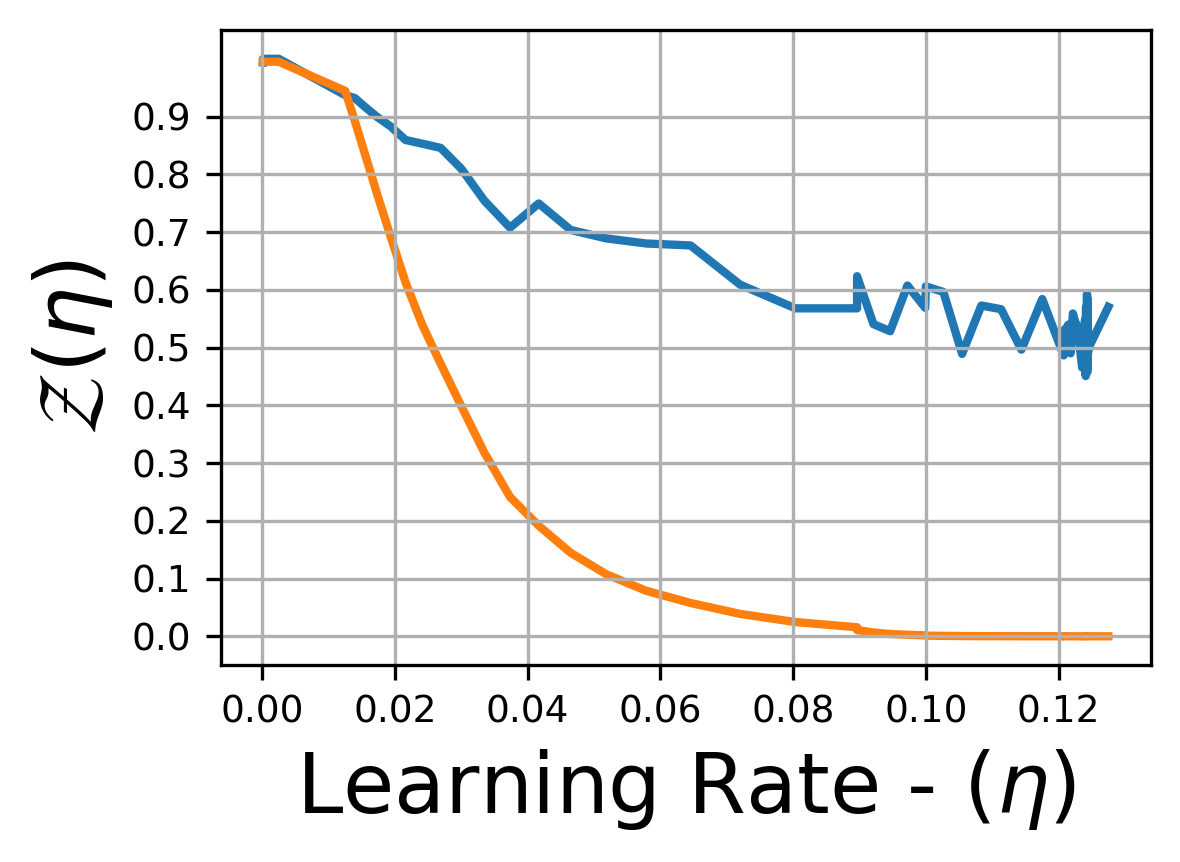}\label{figure_unstable}}
\caption{$\gZ(\GlobalLR)$ (blue) vs. cumprod($\gZ(\GlobalLR)$)\textsuperscript{0.8} (orange) for (a) a stable and (b) an unstable architecture on CIFAR10.}
\vspace{-.5cm}
\end{figure}
\clearpage
\section{Additional Experimental Details for Subsection 4.1}
\label{app_exp_setups}
We note the additional configurations for our experimental setups.

\textbf{Datasets:} For CIFAR10 and CIFAR100, we perform random cropping to $32\times32$ and random horizontal flipping on the training images and make no alterations to the test set. For TinyImageNet, we perform random resized cropping to $64\times64$ and random horizontal flipping on the training images and center crop resizing to $64\times64$ on the test set. For ImageNet, we follow \citet{he2015deep} and perform random resized cropping to $224\times244$ and random horizontal flipping and $256\times256$ resizing with $224\times224$ center cropping on the test set.

\textbf{Additional Configurations:} Experiments on CIFAR10, CIFAR100, and TinyImageNet used mini-batch sizes of 128 and ImageNet experiments used mini-batch sizes  of 256.  For weight decay, $\num{5e-4}$ was used for Adas-variants on CIFAR10 and CIFAR100 experiments and $\num{1e-4}$ for all optimizers on TinyImageNet and ImageNet experiments, with the exception of Adam using a weight decay of $\num{7.8125e-6}$. For Adas-variant, the momentum rate for momentum-SGD was set to $0.9$. All other hyper-parameters for each respective optimizer remained default as reported in their original papers. For author suggested learning rates, for CIFAR10 and CIFAR100, we use the manually tuned suggested learning rates as reported in \citet{wilson2017marginal} for Adam, RMSProp, and AdaGrad. For TinyImageNet and ImageNet, we use the suggested learning rates as reported in each optimizer's respective paper. Refer to Tables \ref{table_lrs_imagenet}-\ref{table_lrs_2d} to see exactly which learning rates were used, as well as the learning rates generated by \AlgName. Further, see \ref{table_lrs_2d} for the learning rates and weight decay reates generated by BO and \AlgName. CIFAR10, CIFAR100, and TinyImageNet experiments were trained for 5 trials with a maximum of 250 epochs and ImageNet experiments were trained for 3 trials with a maximum of 150 epochs. Due to Adas' stable test accuracy behaviour as demonstrated by \citet{hosseini2020adas}, an early-stop criteria, monitoring testing accuracy, was used for CIFAR10, CIFAR100, and ImageNet experiments. For CIFAR10 and CIFAR100, a threshold of $\num{1e-3}$ for Adas\textsuperscript{$\beta=0.8$} and $\num{1e-4}$ for Adas\textsuperscript{$\beta=\{0.9, 0.95\}$} and patience window of $10$ epochs. For ImageNet, a threshold of $\num{1e-4}$ for Adas\textsuperscript{$\beta=\{0.8, 0.9, 0.95\}$} and patience window of $20$ epochs. No early stop is used for Adas\textsuperscript{$\beta=0.975$}.

\textbf{Learning Rates:} We report every learning rate in Tables \ref{table_lrs_imagenet}-\ref{table_lrs_2d}.

\textbf{Random Search:} The search space is set to $[\num{1e-4}, 0.1]$ and a \textit{loguniform} (see \href{https://docs.scipy.org/doc/scipy/reference/generated/scipy.stats.loguniform.html}{SciPy}) distribution is used for sampling. This is motivated by the fact that \AlgName\ also uses and logarithmically-spaced grid space. We note that we ran initial tests against a uniform distribution for sampling was done and showed slightly worse results, as the favouring of smaller learning rates benefits the optimizers we considered. In keeping with \AlgName's design, the learning rate that resulted in highest training accuracy after $5$ epochs was chosen. One could also track testing loss, however we found very little to no differences between the two in initial testing. Further work could include completing both testing loss and testing accuracy baselines, and picking the best one, however this is double the computational that \AlgName\ requires and therefore we deemed it not a fair comparison. Note also we used testing accuracy and not validation accuracy as is normally done, however this only benefits Random Search.

\textbf{Bayesian Optimization:} We used Facebook's \href{https://ax.dev/}{Adaptive Experimentation Platform (AX)} to perform the Bayesian Optimization. In the background, AX uses \citet{balandat2020botorch}, and we refer the reader to that paper for specific details. In keeping with Random Search as well as tutorials on the AX website, testing accuracy was used. Note also we used testing accuracy and not validation accuracy as is normally done, however this only benefits Bayesian Optimization.

\begin{table}[htp]
\centering

\caption{Learning Rates for ResNet34 applied to ImageNet}
\vspace{0.2cm}
\label{table_exp_setup_sugg_r34}
	\footnotesize{
\begin{tabular}{c|c|c}
		\hlinewd{1.3pt}
        Optimizer&Author&\AlgName\\
        \hline
        \hline
        Adam&$0.001$&$0.0001965$\\
        Adas\textsuperscript{($0.9$)}&$0.02$&$0.011479$\\
        Adas\textsuperscript{($0.95$)}&$0.02$&$0.011479$\\
        Adas\textsuperscript{($0.975$)}&$0.02$&$0.011479$\\
        \hlinewd{1.3pt}
		\end{tabular}
	}
	\label{table_lrs_imagenet}
\end{table}

\begin{table}[htp]
    \centering
    \setlength\tabcolsep{2pt}
    \caption{Learning rates for ResNet34 applied to TinyImageNet. These are the one-dimensional hyper-parameter comparison values.}
    \begin{tabular}{c|ccc|c}
        \hlinewd{1.3pt}
        Optimizer&Author&RS&BO&autoHyper\\
        \hline
        \hline
    AdaBound&1.00e-3&1.24e-4&1.64e-4&9.44e-5\\
    AdaGrad&1.00e-2&7.15e-4&7.97e-4&2.24e-3\\
    Adam&1.00e-3&1.75e-4&1.81e-4&1.96e-4\\
    Adas\textsuperscript{$0.9$}&3.00e-2&3.95e-2&6.76e-2&8.59e-3\\
    \hline
    Adas\textsuperscript{$0.8$}&3.00e-2&-&-&1.01e-2\\
    Adas\textsuperscript{$0.95$}&3.00e-2&-&-&8.59e-3\\
    Adas\textsuperscript{$0.975$}&3.00e-2&-&-&8.59e-3\\
    \hlinewd{1.3pt}
    \end{tabular}
    \label{table_lrs_tiny}
\end{table}

\begin{table}[htp]
    \centering
    \setlength\tabcolsep{2pt}
    \caption{Learning rates for various networks applied to CIFAR10 in the one-dimensional comparison.}
    \begin{tabular}{c|ccc|c}
    \multicolumn{5}{c}{(a) ResNet18}\\
    \hlinewd{1.3pt}
    Optimizer&Author&RS&BO&autoHyper\\
    \hline
    \hline
    AdaBound&1.00e-3&2.65e-4&2.55e-4&3.60e-4\\
    AdaGrad&1.00e-2&2.13e-3&2.56e-3&4.97e-3\\
    Adam&3.00e-4&1.45e-4&3.37e-4&6.76e-4\\
    Adas\textsuperscript{$0.9$}&3.00e-2&2.23e-2&2.60e-2&1.04e-2\\
    \hline
    Adas\textsuperscript{$0.8$}&3.00e-2&-&-&1.27e-2\\
    Adas\textsuperscript{$0.95$}&3.00e-2&-&-&1.04e-2\\
    Adas\textsuperscript{$0.975$}&3.00e-2&-&-&1.04e-2\\
    RMSProp&3.00e-4&-&-&4.70e-4\\
    SLS&1.0&-&-&3.42e-2\\
    \hlinewd{1.3pt}
    \multicolumn{5}{c}{}\\
    \multicolumn{5}{c}{(b) ResNet34}\\
    \hlinewd{1.3pt}
    Optimizer&Author&RS&BO&autoHyper\\
    \hline
    \hline
    AdaBound&1.00e-3&3.92e-4&5.47e-4&3.47e-4\\
    AdaGrad&1.00e-2&1.60e-3&1.63e-3&2.86e-3\\
    Adam&3.00e-4&3.30e-4&3.42e-4&3.34e-4\\
    Adas\textsuperscript{$0.9$}&3.00e-2&6.78e-3&4.32e-3&1.24e-2\\
    \hline
    Adas\textsuperscript{$0.8$}&3.00e-2&-&-&1.24e-2\\
    Adas\textsuperscript{$0.95$}&3.00e-2&-&-&1.24e-2\\
    Adas\textsuperscript{$0.975$}&3.00e-2&-&-&1.24e-2\\
    RMSProp&3.00e-4&-&-&1.68e-4\\
    SLS&1.0&-&-&3.42e-2\\
    \hlinewd{1.3pt}
    \multicolumn{5}{c}{}\\
    \multicolumn{5}{c}{(c) ResNeXt50}\\
    \hlinewd{1.3pt}
    Optimizer&Author&RS&BO&autoHyper\\
    \hline
    \hline
    AdaBound&1.00e-3&1.87e-4&4.79e-4&9.72e-4\\
    AdaGrad&1.00e-2&3.80e-3&4.96e-3&8.97e-3\\
    Adam&3.00e-4&1.48e-4&2.82e-4&9.72e-4\\
    Adas\textsuperscript{$0.9$}&3.00e-2&1.38e-2&1.86e-2&2.32e-2\\
    \hline
    Adas\textsuperscript{$0.8$}&3.00e-2&-&-&1.27e-2\\
    Adas\textsuperscript{$0.95$}&3.00e-2&-&-&2.31e-2\\
    Adas\textsuperscript{$0.975$}&3.00e-2&-&-&1.04e-2\\
    RMSProp&3.00e-4&-&-&4.70e-4\\
    SLS&1.0&-&-&3.42e-2\\
    \hlinewd{1.3pt}
    \multicolumn{5}{c}{}\\
    \multicolumn{5}{c}{(d) DenseNet121}\\
    \hlinewd{1.3pt}
    Optimizer&Author&RS&BO&autoHyper\\
    \hline
    \hline
    AdaBound&1.00e-3&8.91e-4&9.21e-4&3.02e-3\\
    AdaGrad&1.00e-2&4.85e-3&8.37e-3&1.54e-2\\
    Adam&3.00e-4&7.50e-4&4.81e-4&2.01e-3\\
    Adas\textsuperscript{$0.9$}&3.00e-2&2.09e-2&3.07e-2&5.98e-2\\
    \hline
    Adas\textsuperscript{$0.8$}&3.00e-2&-&-&6.10e-2\\
    Adas\textsuperscript{$0.95$}&3.00e-2&-&-&4.92e-2\\
    Adas\textsuperscript{$0.975$}&3.00e-2&-&-&5.98e-2\\
    RMSProp&3.00e-4&-&-&2.01e-3\\
    SLS&1.0&-&-&3.12e-3\\
    \hlinewd{1.3pt}
    \end{tabular}
    \label{table_lrs_c10}
\end{table}

\begin{table}[htp]
    \centering
    \setlength\tabcolsep{2pt}
    \caption{Learning rates for various networks applied to CIFAR100 in the one-dimensional comparison.}
    \begin{tabular}{c|ccc|c}
    \multicolumn{5}{c}{(a) ResNet18}\\
    \hlinewd{1.3pt}
    Optimizer&Author&RS&BO&autoHyper\\
    \hline
    \hline
    AdaBound&1.00e-3&3.43e-4&2.92e-4&2.49e-4\\
    AdaGrad&1.00e-2&4.51e-3&2.87e-3&4.97e-3\\
    Adam&3.00e-4&3.59e-4&2.87e-4&6.76e-4\\
    Adas\textsuperscript{$0.9$}&3.00e-2&5.52e-2&3.06e-2&1.27e-2\\
    \hline
    Adas\textsuperscript{$0.8$}&3.00e-2&-&-&1.27e-2\\
    Adas\textsuperscript{$0.95$}&3.00e-2&-&-&1.04e-2\\
    Adas\textsuperscript{$0.975$}&3.00e-2&-&-&7.07e-3\\
    RMSProp&3.00e-4&-&-&4.70e-4\\
    SLS&1.0&-&-&3.42e-2\\
    \hlinewd{1.3pt}
    \multicolumn{5}{c}{}\\
    \multicolumn{5}{c}{(b) ResNet34}\\
    \hlinewd{1.3pt}
    Optimizer&Author&RS&BO&autoHyper\\
    \hline
    \hline
    AdaBound&1.00e-3&3.53e-4&1.76e-4&3.47e-4\\
    AdaGrad&1.00e-2&1.83e-3&2.35e-3&2.24e-3\\
    Adam&3.00e-4&1.25e-4&2.42e-4&2.41e-4\\
    Adas\textsuperscript{$0.9$}&3.00e-2&9.25e-3&2.14e-3&1.02e-2\\
    \hline
    Adas\textsuperscript{$0.8$}&3.00e-2&-&-&1.03e-2\\
    Adas\textsuperscript{$0.95$}&3.00e-2&-&-&1.50e-2\\
    Adas\textsuperscript{$0.975$}&3.00e-2&-&-&1.02e-2\\
    RMSProp&3.00e-4&-&-&1.97e-4\\
    SLS&1.0&-&-&3.42e-2\\
    \hlinewd{1.3pt}
    \multicolumn{5}{c}{}\\
    \multicolumn{5}{c}{(c) ResNeXt50}\\
    \hlinewd{1.3pt}
    Optimizer&Author&RS&BO&autoHyper\\
    \hline
    \hline
    AdaBound&1.00e-3&3.69e-4&5.03e-4&9.72e-4\\
    AdaGrad&1.00e-2&2.74e-3&5.86e-3&1.19e-2\\
    Adam&3.00e-4&4.38e-4&5.44e-4&9.72e-4\\
    Adas\textsuperscript{$0.9$}&3.00e-2&1.16e-2&1.26e-2&2.75e-2\\
    \hline
    Adas\textsuperscript{$0.8$}&3.00e-2&-&-&1.27e-2\\
    Adas\textsuperscript{$0.95$}&3.00e-2&-&-&2.27e-2\\
    Adas\textsuperscript{$0.975$}&3.00e-2&-&-&7.0e-3\\
    RMSProp&3.00e-4&-&-&4.70e-4\\
    SLS&1.0&-&-&3.42e-2\\
    \hlinewd{1.3pt}
    \multicolumn{5}{c}{}\\
    \multicolumn{5}{c}{(d) DenseNet121}\\
    \hlinewd{1.3pt}
    Optimizer&Author&RS&BO&autoHyper\\
    \hline
    \hline
    AdaBound&1.00e-3&8.91e-4&9.21e-4&3.02e-3\\
    AdaGrad&1.00e-2&4.85e-3&8.37e-3&1.54e-2\\
    Adam&3.00e-4&7.50e-4&4.81e-4&2.01e-3\\
    Adas\textsuperscript{$0.9$}&3.00e-2&2.09e-2&3.07e-2&5.98e-2\\
    \hline
    Adas\textsuperscript{$0.8$}&3.00e-2&-&-&3.98e-2\\
    Adas\textsuperscript{$0.95$}&3.00e-2&-&-&3.57e-2\\
    Adas\textsuperscript{$0.975$}&3.00e-2&-&-&5.04e-2\\
    RMSProp&3.00e-4&-&-&6.76e-2\\
    SLS&1.0&-&-&8.79e-2\\
    \hlinewd{1.3pt}
    \end{tabular}
    \label{table_lrs_c100}
\end{table}

\begin{table}[htp]
    \centering
    \setlength\tabcolsep{2pt}
    \caption{Learning rates for ResNet34 applied to various dataset for the two-dimensional comparison.}
    \begin{tabular}{c|cc|cc}
    \multicolumn{5}{c}{(a) TinyImageNet}\\
    \hlinewd{1.3pt}
    Optimizer&\multicolumn{2}{c}{BO}&\multicolumn{2}{c}{autoHyper}\\
    \hline
    &LR ($\eta$)&WD ($\gamma$)&LR ($\eta$)&WD ($\gamma$)\\
    \hline
    \hline
    AdaBound&5.33e-5&6.28e-4&1.62e-5&7.17e-8\\
    AdaGrad&5.21e-4&1.27e-3&4.27e-3&1.04e-7\\
    Adam&7.03e-4&5.50e-6&2.19e-4&1.00e-7\\
    Adas\textsuperscript{$0.9$}&8.99e-3&1.05e-6&1.89e-2&9.67e-7\\
    \hlinewd{1.3pt}
    \multicolumn{5}{c}{}\\
    \multicolumn{5}{c}{(b) CIFAR10}\\
    \hlinewd{1.3pt}
    Optimizer&\multicolumn{2}{c}{BO}&\multicolumn{2}{c}{autoHyper}\\
    \hline
    &LR ($\eta$)&WD ($\gamma$)&LR ($\eta$)&WD ($\gamma$)\\
    \hline
    \hline
    AdaBound&9.66e-6&4.32e-6&6.67e-4&3.41e-8\\
    AdaGrad&2.92e-3&3.29e-5&6.20e-3&3.17e-7\\
    Adam&4.35e-4&1.68e-8&6.67e-4&1.00e-7\\
    Adas\textsuperscript{$0.9$}&2.86e-3&1.87e-6&2.74e-2&6.43e-7\\

    \hlinewd{1.3pt}
    \multicolumn{5}{c}{}\\
    \multicolumn{5}{c}{(c) CIFAR100}\\
    \hlinewd{1.3pt}
    Optimizer&\multicolumn{2}{c}{BO}&\multicolumn{2}{c}{autoHyper}\\
    \hline
    &LR ($\eta$)&WD ($\gamma$)&LR ($\eta$)&WD ($\gamma$)\\
    \hline
    \hline
    AdaBound&1.19e-5&5.73e-7&3.67e-6&2.35e-8\\
    AdaGrad&4.08e-3&1.03e-3&6.20e-3&1.51e-7\\
    Adam&1.52e-3&1.01e-6&4.60e-4&7.17e-8\\
    Adas\textsuperscript{$0.9$}&7.70e-3&1.12e-7&2.74e-2&4.60e-7\\
    \hlinewd{1.3pt}
    \end{tabular}
    \label{table_lrs_2d}
\end{table}

\clearpage
\section{Additional Results for Subsection 4.2}
\label{app_results}
\begin{table}[h!]
    \setlength\tabcolsep{1pt} 
    \center
	\caption{Final epoch (250) top-1 test accuracies average over each trial for one-dimensional search ($\lambda = \eta$). Values marked with a `*' indicate early-stopping. The best result is highlighted in {\color{best}green}, and for \AlgName\ results, {\color{close}orange} highlights when the results lie with the standard deviation from the best.}
	\label{table_add_results}
	\footnotesize{
		\begin{tabular}{cccc|c}
		\multicolumn{5}{c}{(a) ResNet34 on TinyImageNet}\\
		\hlinewd{1pt}
		Optimizer&Author&RS&BO&\AlgName\\
		\hline
		\hline
		Adas\textsuperscript{$0.8$}&$57.98_{\pm0.44}$&-&-&${\color{best}\bm{58.02_{\pm0.42}}}$\\
		Adas\textsuperscript{$0.95$}&$60.74_{\pm0.20}$&-&-&${\color{best}\bm{62.28_{\pm0.44}}}$\\
		Adas\textsuperscript{$0.975$}&$61.44_{\pm0.27}$&-&-&${\color{best}\bm{61.81_{\pm0.45}}}$\\

		\hlinewd{1pt}
        \\
        \multicolumn{5}{c}{(c) ResNet34 on CIFAR10}\\
		\hlinewd{1pt}
		Optimizer&Author&RS&BO&\AlgName\\
		\hline
		\hline
		Adas\textsuperscript{$0.8$}&$93.02_{\pm0.13}$&-&-&${\color{best}\bm{93.40^*_{\pm0.15}}}$\\
		Adas\textsuperscript{$0.95$}&${\color{best}\bm{95.20_{\pm0.11}}}$&-&-&${\color{close}\bm{95.08^*_{\pm0.18}}}$\\
		Adas\textsuperscript{$0.975$}&${\color{best}\bm{95.24_{\pm0.15}}}$&-&-&${\color{close}\bm{95.13_{\pm0.11}}}$\\
		RMSProp&$92.90_{\pm0.29}$&-&-&${\color{best}\bm{93.03_{\pm0.23}}}$\\
		SLS&${\color{best}\bm{93.45_{\pm0.16}}}$&-&-&${\color{close}\bm{93.33_{\pm0.06}}}$\\

		\hlinewd{1pt}
        \\
        \multicolumn{5}{c}{(d) ResNet34 on CIFAR100}\\
		\hlinewd{1pt}
		Optimizer&Author&RS&BO&\AlgName\\
		\hline
		\hline
		Adas\textsuperscript{$0.8$}&${\color{best}\bm{74.21_{\pm0.26}}}$&-&-&$73.58^*_{\pm0.36}$\\
		Adas\textsuperscript{$0.95$}&${\color{best}\bm{77.60_{\pm0.22}}}$&-&-&${\color{close}\bm{77.48^*_{\pm0.37}}}$\\
		Adas\textsuperscript{$0.975$}&$78.00_{\pm0.28}$&-&-&${\color{best}\bm{78.26_{\pm0.35}}}$\\
		RMSProp&$70.25_{\pm0.29}$&-&-&${\color{best}\bm{70.57_{\pm0.40}}}$\\
		SLS&$73.22_{\pm0.11}$&-&-&${\color{best}\bm{73.77_{\pm0.12}}}$\\
		\hlinewd{1pt}
        \\
		\multicolumn{5}{c}{(e) ResNet18 on CIFAR10}\\
		\hlinewd{1pt}
		Optimizer&Author&RS&BO&\AlgName\\
		\hline
		\hline
		AdaBound&$92.35_{\pm0.18}$&$92.64_{\pm0.21}$&${\color{best}{\color{best}\bm{93.15_{\pm0.11}}}}$&$92.85_{\pm0.06}$\\
        AdaGrad&${\color{best}\bm{91.23_{\pm0.25}}}$&$89.71_{\pm0.18}$&$90.00_{\pm0.08}$&${\color{close}\bm{90.87_{\pm0.14}}}$\\
        Adam&$92.93_{\pm0.22}$&$92.59_{\pm0.05}$&$92.92_{\pm0.18}$&${\color{best}\bm{92.95_{\pm0.24}}}$\\
        Adas\textsuperscript{$0.9$}&${\color{best}\bm{94.05_{\pm0.10}}}$&$94.11_{\pm0.13}$&$94.01_{\pm0.05}$&$93.75^*_{\pm0.12}$\\
        \hline
        \hline
        Adas\textsuperscript{$0.8$}&${\color{best}\bm{92.92_{\pm0.19}}}$&-&-&${\color{close}\bm{92.80^*_{\pm0.16}}}$\\
        Adas\textsuperscript{$0.95$}&${\color{best}\bm{94.93_{\pm0.11}}}$&-&-&${\color{close}\bm{94.74^*_{\pm0.16}}}$\\
        Adas\textsuperscript{$0.975$}&${\color{best}\bm{95.14_{\pm0.20}}}$&-&-&${\color{close}\bm{94.94_{\pm0.04}}}$\\
        RMSProp&$92.62_{\pm0.30}$&-&-&${\color{best}\bm{92.69_{\pm0.33}}}$\\
        SLS&${\color{best}\bm{93.45_{\pm0.16}}}$&-&-&${\color{close}\bm{93.33_{\pm0.06}}}$\\

		\hlinewd{1pt}
        \\
        \multicolumn{5}{c}{(f) ResNet18 on CIFAR100}\\
		\hlinewd{1pt}
		Optimizer&Author&RS&BO&\AlgName\\
		\hline
		\hline
		 Adas\textsuperscript{$0.8$}&${\color{best}\bm{73.59_{\pm0.09}}}$&-&-&${\color{close}\bm{73.38^*_{\pm0.28}}}$\\
        Adas\textsuperscript{$0.95$}&${\color{best}\bm{76.53_{\pm0.30}}}$&-&-&${\color{close}\bm{76.49^*_{\pm0.37}}}$\\
        Adas\textsuperscript{$0.975$}&${\color{best}\bm{77.23_{\pm0.09}}}$&-&-&$76.68_{\pm0.18}$\\
        RMSProp&${\color{best}\bm{70.08_{\pm0.23}}}$&-&-&${\color{close}\bm{69.28_{\pm0.50}}}$\\
        SLS&$73.22_{\pm0.11}$&-&-&${\color{best}\bm{73.77_{\pm0.12}}}$\\

		\hlinewd{1pt}
		\end{tabular}
	}
\end{table}
\begin{table}[h!]
    \setlength\tabcolsep{1pt} 
    \center
	\caption{Final epoch (250) top-1 test accuracies average over each trial for one-dimensional search ($\lambda = \eta$).  Values marked with a `*' indicate early-stopping. The best result is highlighted in {\color{best}green}, and for \AlgName\ results, {\color{close}orange} highlights when the results lie with the standard deviation from the best.}
	\label{table_add_results_2}
	\footnotesize{
		\begin{tabular}{cccc|c}
		\multicolumn{5}{c}{(g) ResNeXt50 on CIFAR10}\\
		\hlinewd{1pt}
		Optimizer&Author&RS&BO&\AlgName\\
		\hline
		\hline
		AdaBound&$91.42_{\pm0.42}$&${\color{best}\bm{92.37_{\pm0.19}}}$&$92.10_{\pm0.19}$&$91.69_{\pm0.33}$\\
        AdaGrad&$90.07_{\pm0.27}$&$89.02_{\pm0.23}$&$89.26_{\pm0.26}$&${\color{best}\bm{90.13_{\pm0.19}}}$\\
        Adam&$92.18_{\pm0.31}$&$91.90_{\pm0.17}$&${\color{best}\bm{92.29_{\pm0.33}}}$&${\color{close}\bm{92.12_{\pm0.07}}}$\\
        Adas\textsuperscript{$0.9$}&${\color{best}\bm{93.60_{\pm0.16}}}$&$93.51_{\pm0.12}$&$93.39_{\pm0.12}$&${\color{close}\bm{93.51^*_{\pm0.12}}}$\\
        \hline
        \hline
        Adas\textsuperscript{$0.8$}&${\color{best}\bm{91.56_{\pm0.07}}}$&-&-&${\color{close}\bm{91.49^*_{\pm0.16}}}$\\
        Adas\textsuperscript{$0.95$}&${\color{best}\bm{94.62_{\pm0.10}}}$&-&-&${\color{close}\bm{94.61^*_{\pm0.11}}}$\\
        Adas\textsuperscript{$0.975$}&${\color{best}\bm{95.03_{\pm0.12}}}$&-&-&${\color{close}\bm{95.02_{\pm0.06}}}$\\
        RMSProp&${\color{best}\bm{92.15_{\pm0.20}}}$&-&-&$91.34_{\pm0.59}$\\
        SLS&$93.49_{\pm0.14}$&-&-&${\color{best}\bm{93.56_{\pm0.20}}}$\\
		\hlinewd{1pt}
        \\
		\multicolumn{5}{c}{(h) ResNeXt50 on CIFAR100}\\
		\hlinewd{1pt}
		Optimizer&Author&RS&BO&\AlgName\\
		\hline
		\hline
		AdaBound&$71.43_{\pm0.30}$&${\color{best}\bm{72.50_{\pm0.28}}}$&$72.27_{\pm0.30}$&${\color{close}\bm{71.20_{\pm0.34}}}$\\
        AdaGrad&$65.66_{\pm0.36}$&$62.32_{\pm0.15}$&${\color{best}\bm{66.07_{\pm0.38}}}$&${\color{close}\bm{66.03_{\pm0.56}}}$\\
        Adam&$70.32_{\pm0.46}$&${\color{best}\bm{70.33_{\pm0.36}}}$&$69.95_{\pm0.40}$&$69.12_{\pm0.16}$\\
        Adas\textsuperscript{$0.9$}&${\color{best}\bm{74.43_{\pm0.14}}}$&$73.75_{\pm0.30}$&$73.97_{\pm0.16}$&${\color{close}\bm{74.41^*_{\pm0.26}}}$\\
        \hline
        \hline
        Adas\textsuperscript{$0.8$}&${\color{best}\bm{72.41_{\pm0.16}}}$&-&-&${\color{close}\bm{72.00^*_{\pm0.44}}}$\\
        Adas\textsuperscript{$0.95$}&${\color{best}\bm{75.95_{\pm0.26}}}$&-&-&${\color{close}\bm{75.63^*_{\pm0.12}}}$\\
        Adas\textsuperscript{$0.975$}&$76.46_{\pm0.24}$&-&-&${\color{best}\bm{76.58_{\pm0.21}}}$\\
        RMSProp&${\color{best}\bm{69.45_{\pm1.17}}}$&-&-&$67.17_{\pm0.70}$\\
        SLS&${\color{best}\bm{72.08_{\pm0.43}}}$&-&-&${\color{close}\bm{71.82_{\pm0.22}}}$\\

		\hlinewd{1pt}
		\\
		\multicolumn{5}{c}{(i) DenseNet121 on CIFAR10}\\
		\hlinewd{1pt}
		Optimizer&Author&RS&BO&\AlgName\\
		\hline
		\hline
		 Adas\textsuperscript{$0.8$}&$91.28_{\pm0.23}$&-&-&${\color{best}\bm{91.59^*_{\pm0.25}}}$\\
        Adas\textsuperscript{$0.95$}&${\color{best}\bm{93.51_{\pm0.20}}}$&-&-&${\color{close}\bm{93.33^*_{\pm0.24}}}$\\
        Adas\textsuperscript{$0.975$}&${\color{best}\bm{93.83_{\pm0.20}}}$&-&-&${\color{best}\bm{93.47_{\pm0.24}}}$\\
        RMSProp&$91.29_{\pm0.20}$&-&-&${\color{best}\bm{91.83_{\pm0.30}}}$\\
        SLS&$93.16_{\pm0.13}$&-&-&${\color{best}\bm{93.36_{\pm0.18}}}$\\

		\hlinewd{1pt}
        \\
        \multicolumn{5}{c}{(j) DenseNet121 on CIFAR100}\\
		\hlinewd{1pt}
		Optimizer&Author&RS&BO&\AlgName\\
		\hline
		\hline
		 Adas\textsuperscript{$0.8$}&$70.63_{\pm0.33}$&-&-&${\color{best}\bm{71.01^*_{\pm0.28}}}$\\
        Adas\textsuperscript{$0.95$}&${\color{best}\bm{74.22_{\pm0.24}}}$&-&-&${\color{close}\bm{73.98^*_{\pm0.33}}}$\\
        Adas\textsuperscript{$0.975$}&${\color{best}\bm{74.10_{\pm0.47}}}$&-&-&${\color{close}\bm73.97_{\pm0.36}}$\\
        RMSProp&$66.61_{\pm0.58}$&-&-&${\color{best}\bm{68.13_{\pm0.00}}}$\\
        SLS&$69.44_{\pm0.61}$&-&-&${\color{best}\bm{70.25_{\pm0.19}}}$\\

		\hlinewd{1pt}
        \\
		\end{tabular}
	}
\end{table}

\clearpage
\begin{figure*}[!ht]
\begin{center}
\includegraphics[width=\textwidth]{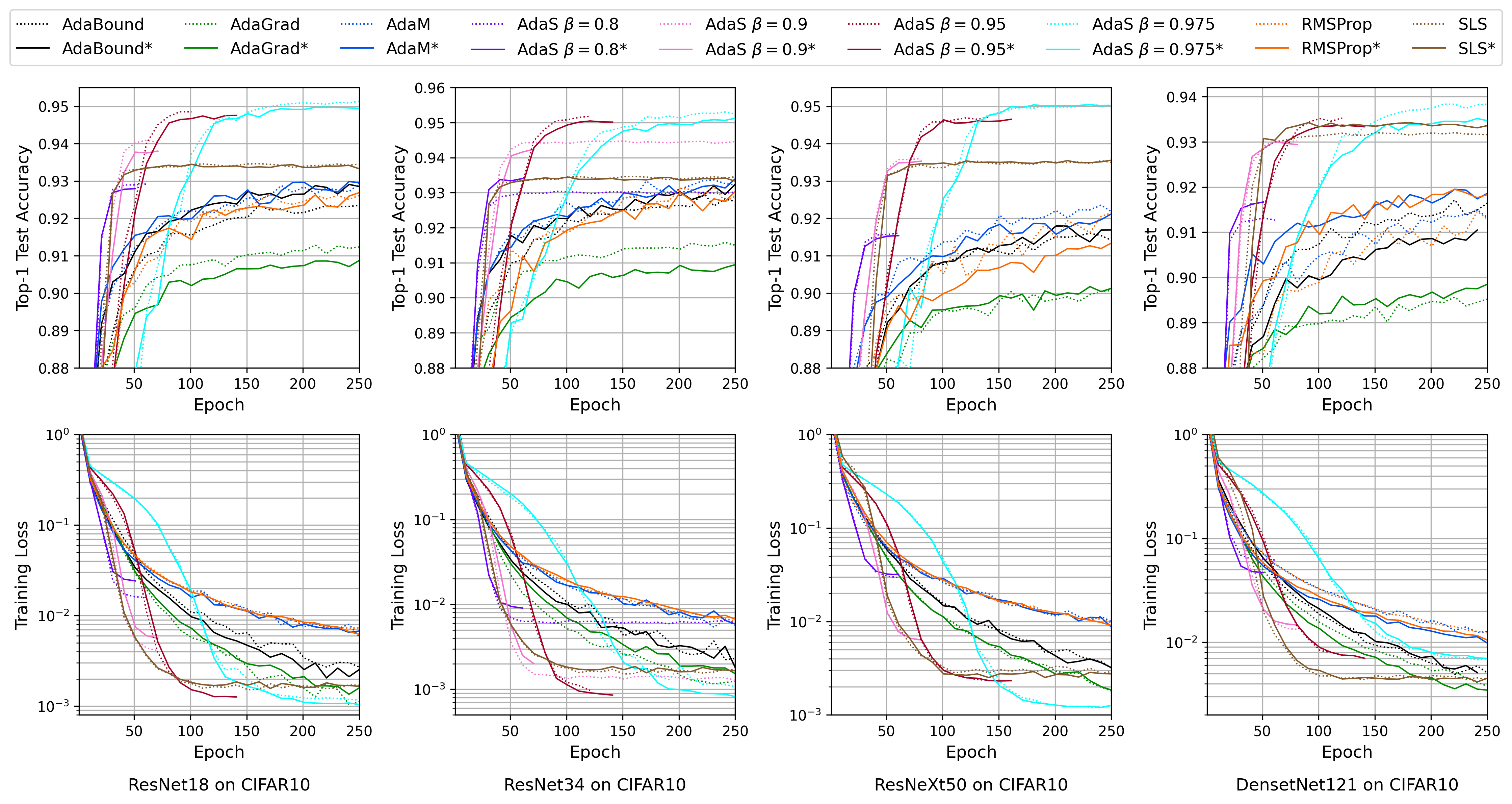}
\includegraphics[width=\textwidth]{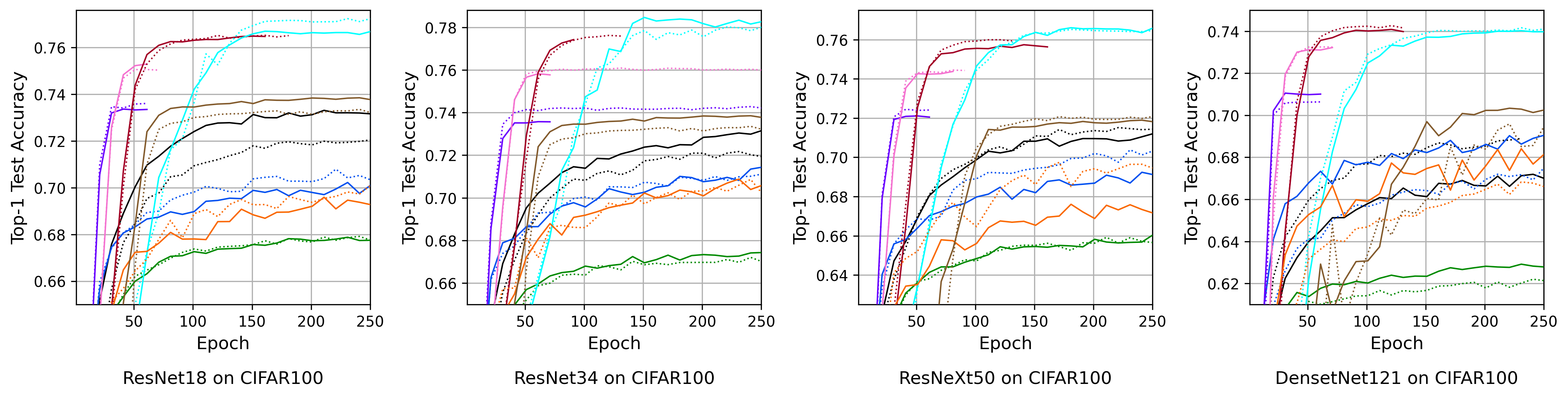}
\subfigure[]{\includegraphics[width=\textwidth]{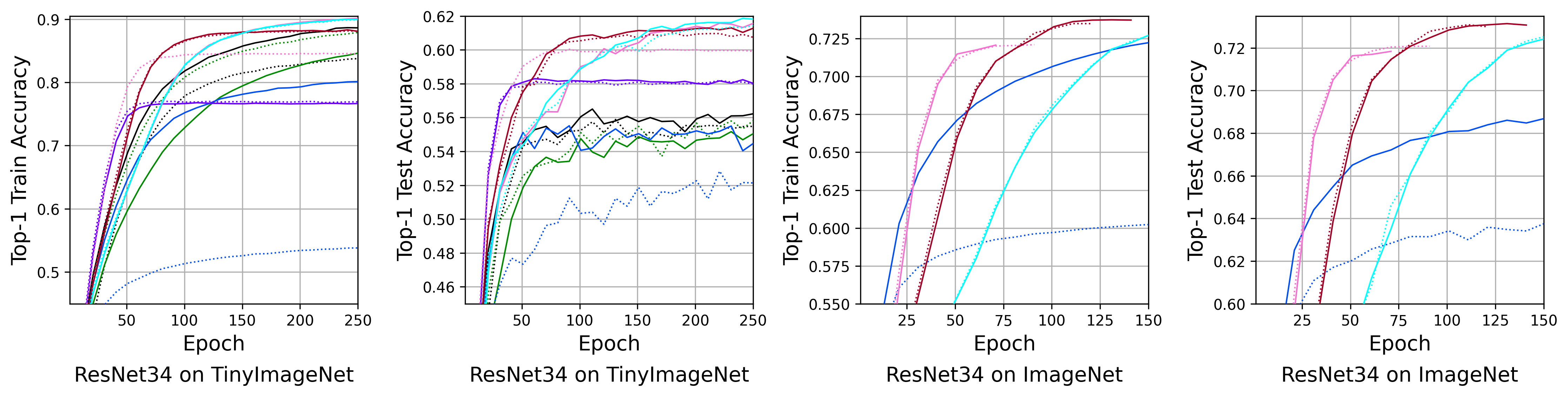}\label{figure_results}}
\end{center}
\caption{{\color{changed}Results of the (a) ablative study and (b) Random Search comparison experiments. Titles below plots indicate what experiment the above plots refers to. Legend labels marked by `*' (solid lines) show results for \AlgName\ generated learning rates and dotted lines are the (a) baselines and (b) Random Search results.}}
\end{figure*}
\begin{figure*}[htp]
\begin{center}
\includegraphics[width=0.8\textwidth]{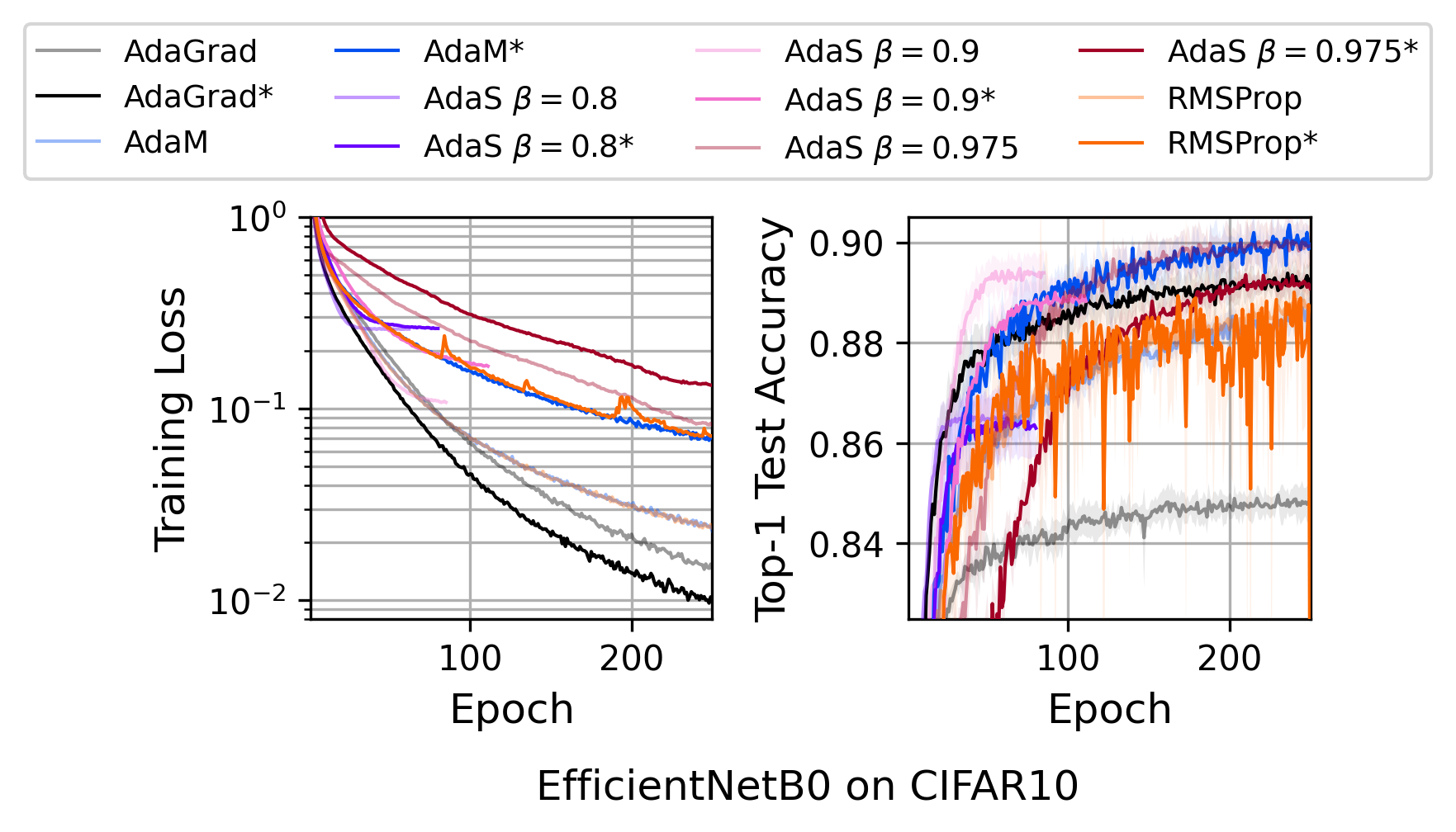}
\end{center}
\caption{Test accuracy and trianing loss for EfficientNetB0 applied to CIFAR100. Importantly, EfficientNetB0 is an unstable network architecture in relation to our response surface and yet our method, \AlgName, is still able to converge and achieve competitive performance.}
\label{eff_results}
\end{figure*}

\begin{figure*}[t]
\begin{center}
\includegraphics[width=0.8\textwidth]{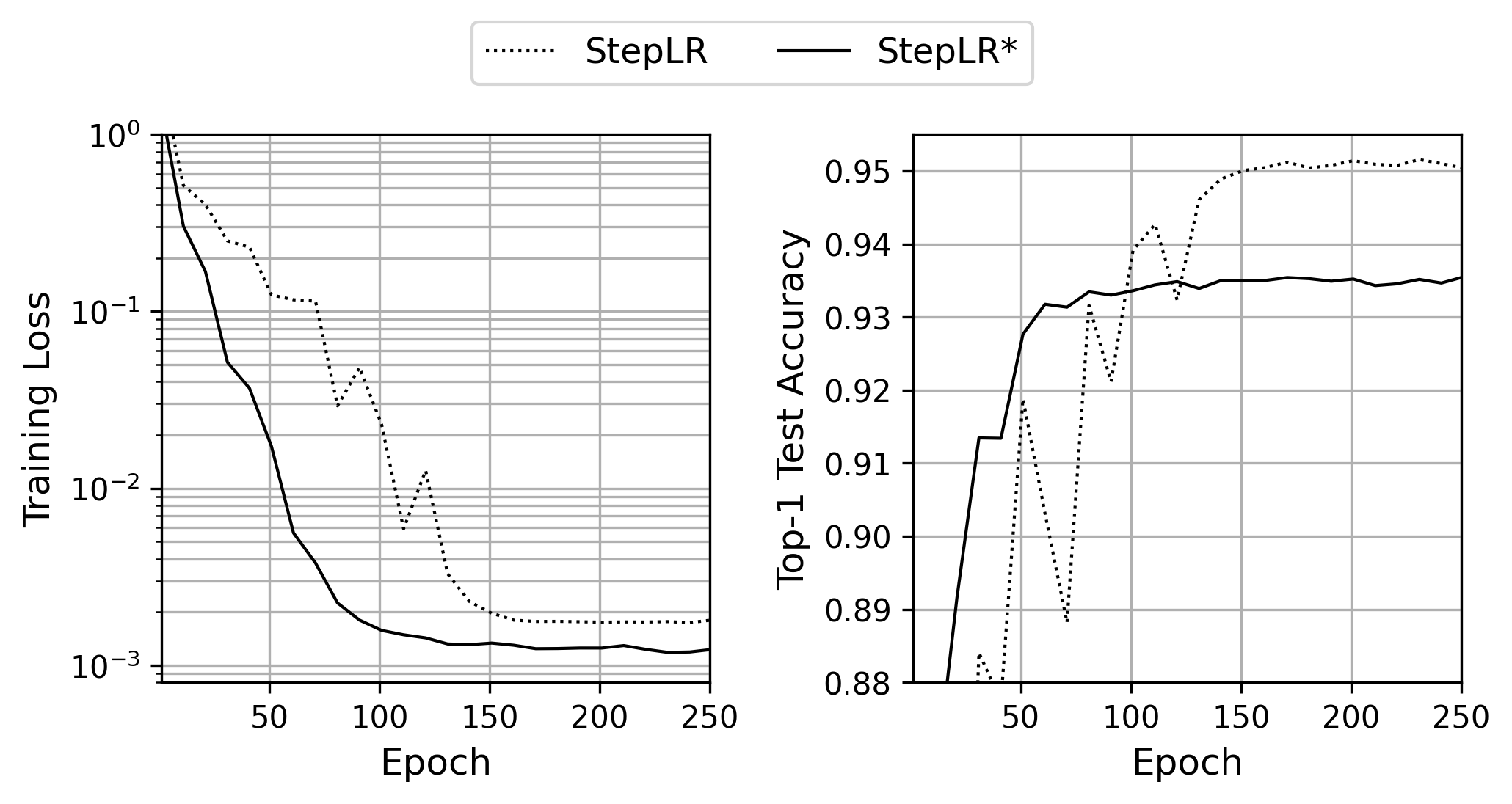}
\end{center}
\caption{{\color{changed}Demonstration of the importance of initial learning rate in scheduled learning rate case, for ResNet18 applied on CIFAR10, using Step-Decay method with step-size = 25 epochs and decay rate = 0.5. As before, the dotted line represents the baseline results, with initial learning rate = 0.1, and the solid line represents the results using autoHyper's suggested learning rate of 0.008585. These results highlight the importance of initial learning rate, even when using a scheduled learning rate heuristic, and demonstrates the importance of the additional step-size and decay rate hyper-parameters. Despite better initial performance from the autoHyper suggest learning rate, the step-size and decay rate choice cause the performance to plateau too early.}}
\label{steplr_results}
\end{figure*}

\begin{figure*}[t]
\begin{center}
\includegraphics[width=\textwidth]{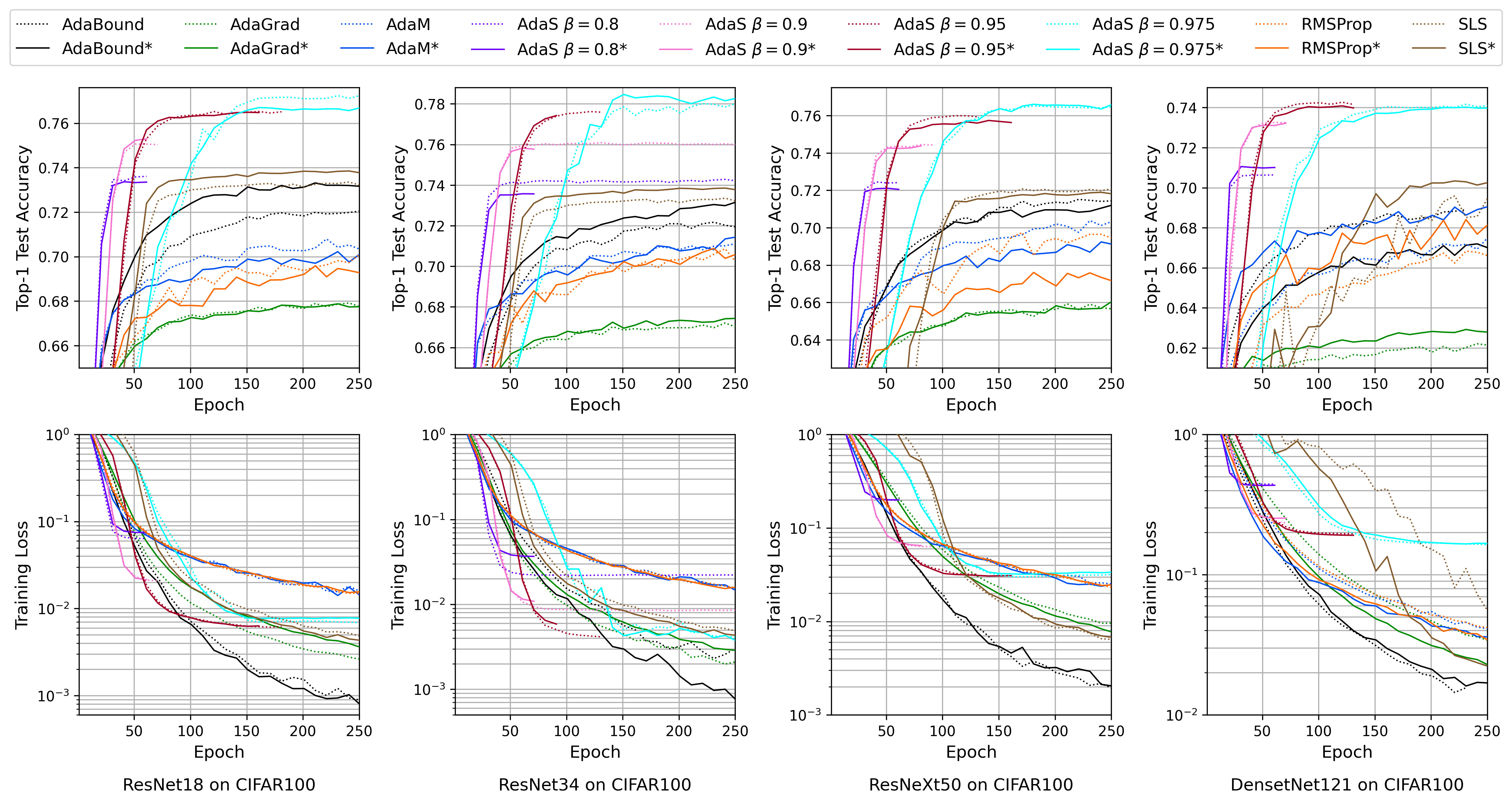}
\includegraphics[width=\textwidth]{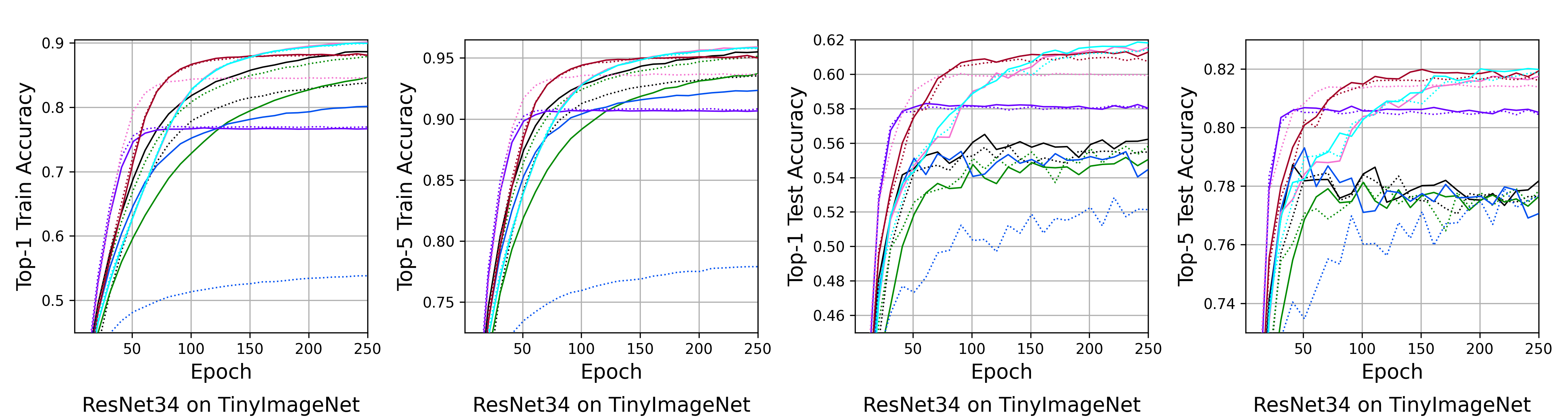}
\includegraphics[width=\textwidth]{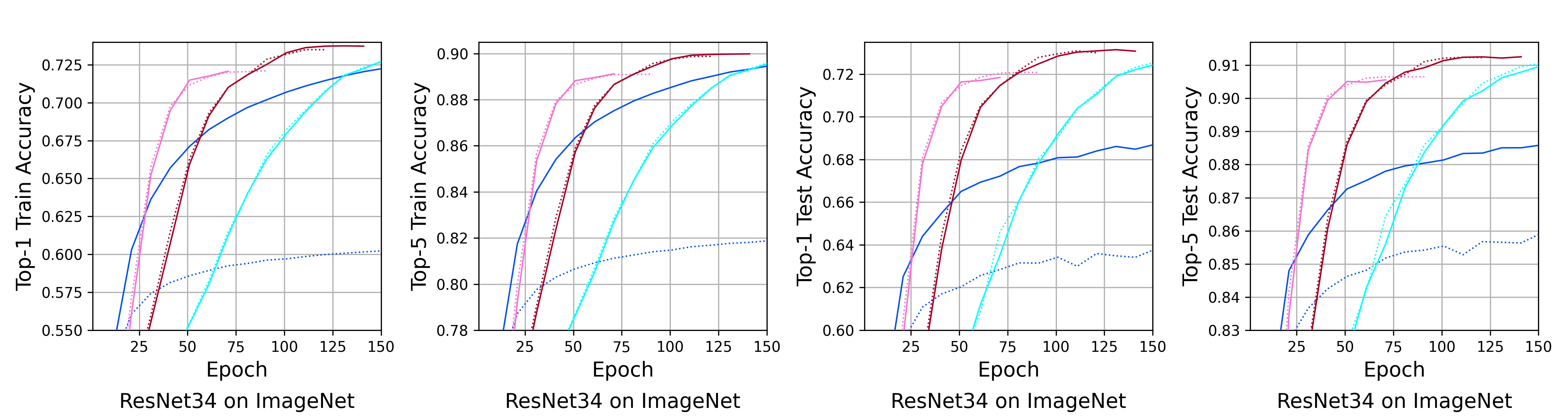}
\end{center}
\caption{Full results of CIFAR100, TinyImageNet, and ImageNet experiments. Top-1 test accuracy and training losses are reported for CIFAR100 experiments and top-1 and top-5 test and training accuracies are reported for TinyImageNet and ImageNet. Titles below the figures indicate to which experiments the above figures belong to. As before, lines indicated by the `*' (solid lines), are results using initial learning rate as suggested by autoHyper.} 
\label{add_results}
\end{figure*}

\begin{figure*}[t]
\begin{center}
\includegraphics[width=\textwidth]{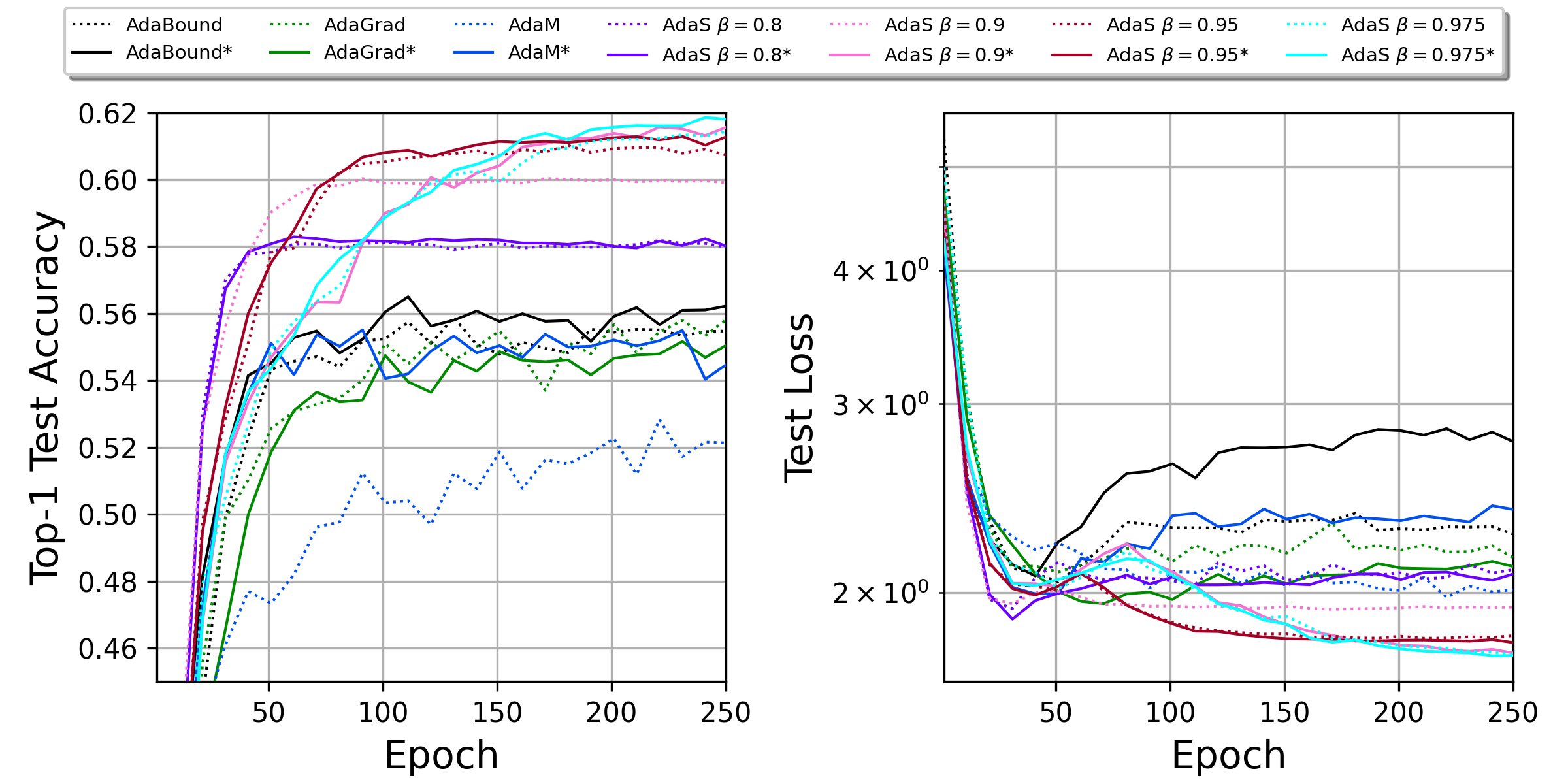}
\end{center}
\caption{{\color{changed}Top-1 Test Accuracy and Test Loss for ResNet34 Experiments applied on TinyImageNet. As before, lines indicated by the `*' (solid lines), are results using initial learning rate as suggested by autoHyper. These results visualize the inconsistency in tracking test loss as a metric to optimize final testing accuracy. This can be seen, for example, when looking at the test loss and test accuracy plots for Adam, where the test loss for the baseline is lower than that of the \AlgName\ suggested results but \AlgName\ achieves better test accuracy. These results also highlight the instability of tracking testing accuracy or less instead of the metric defined in Equation 5.}} 
\label{add_results_loss}
\end{figure*}

\end{document}